\documentclass[letterpaper]{article} 
\usepackage{aaai24}  
\usepackage{times}  
\usepackage{helvet}  
\usepackage{courier}  
\usepackage[hyphens]{url}  
\usepackage{graphicx} 
\usepackage{natbib}  
\usepackage{caption} 
\usepackage{graphicx}
\usepackage{url}
\usepackage{multirow}
\usepackage{amsmath,amssymb,amsthm}
\usepackage{booktabs}
\usepackage{makecell}
\usepackage{algorithm}
\usepackage{algorithmicx}
\usepackage[noend]{algpseudocode}
\usepackage{newfloat}
\usepackage{sidecap}
\usepackage{dsfont}
\usepackage{placeins}

\theoremstyle{plain}
\newtheorem{theorem}{Theorem}[section]

\newtheorem{lemma}[theorem]{Lemma}

\theoremstyle{definition}

\theoremstyle{remark}
\newtheorem{remark}[theorem]{Remark}

\DeclareMathOperator*{\argmax}{argmax}
\DeclareMathOperator*{\argmin}{argmin}

\usepackage{listings}
\DeclareCaptionStyle{ruled}{labelfont=normalfont,labelsep=colon,strut=off} 
\floatstyle{ruled}
\newfloat{listing}{tb}{lst}{}
\floatname{listing}{Listing}
\makeatletter
\renewcommand{\Function}[2]{%
  \csname ALG@cmd@\ALG@L @Function\endcsname{#1}{#2}%
  \def\jayden@currentfunction{#1}%
}
\newcommand{\funclabel}[1]{%
  \@bsphack
  \protected@write\@auxout{}{%
    \string\newlabel{#1}{{\jayden@currentfunction}{\thepage}}%
  }%
  \@esphack
}
\makeatother

\title{Hard Regularization to Prevent Deep Online Clustering Collapse without Data Augmentation}
\author{
    Louis Mahon\textsuperscript{\rm 1, \rm 3},
    Thomas Lukasiewicz\textsuperscript{\rm 2, \rm 3}
}
\affiliations{
    \textsuperscript{\rm 1}School of Informatics, University of Edinburgh, UK\\
    \textsuperscript{\rm 2}Institute of Logic and Computation, Vienna University of Technology, Austria\\
    \textsuperscript{\rm 3}Department of Computer Science, University of Oxford, UK\\
    louis.mahon@inf.ed.ac.uk
}

\begin{document}
\pagenumbering{arabic}

\maketitle

\begin{abstract}
Online deep clustering refers to the joint use of a feature extraction network and a clustering model to assign cluster labels to each new data point or batch as it is processed. While faster and more versatile than offline methods, online clustering can easily reach the collapsed solution where the encoder maps all inputs to the same point and all are put into a single cluster. Successful existing models have employed various techniques to avoid this problem, most of which require data augmentation or which aim to make the average soft assignment across the dataset the same for each cluster. We propose a method that does not require data augmentation, and that, differently from existing methods, regularizes the hard assignments. Using a Bayesian framework, we derive an intuitive optimization objective that can be straightforwardly included in the training of the encoder network. Tested on four image datasets and one human-activity recognition dataset, it consistently avoids collapse more robustly than other methods and leads to more accurate clustering. We also conduct further experiments and analyses justifying our choice to regularize the hard cluster assignments. Code is available at \url{https://github.com/Lou1sM/online_hard_clustering}.
\end{abstract}

\section{Introduction}

Deep clustering refers to the combination of deep learning and clustering, where the data are first encoded with a deep neural network to a feature space, and then clustering is performed in the feature space. Clustering models can be classified as offline or online. Offline models process the entire dataset, and then assign all cluster labels at once. Online models, in contrast, assign a cluster label to each data point, or each batch, as it is processed. Offline methods can produce more accurate clusterings (e.g., \citet{mahon2021selective,niu2021spice} reaching close to supervised performance), as they can leverage information from later data points when assigning cluster labels to earlier data points. However, they are more expensive to train, as they must alternate between encoding the entire dataset, or training the encoder for some number of epochs, and clustering the encoded data. Online models, on the other hand, can jointly encode and cluster, so are less computationally expensive. They are also more versatile, being applicable in real-world settings where new data is constantly becoming available, as opposed to offline methods, which are limited to fixed pre-defined datasets \cite{silva2013data}.

The disadvantage of online methods is that they are more difficult to train. In particular, they run the risk of producing a degenerate solution where the large majority of data points are concentrated in a small number of clusters. In the extreme case, all points are placed into the same cluster. For example, an intuitive way to formulate a training procedure for online deep clustering is to update both the encoder network and the clustering parameters to make the encoding of each point close to its cluster centroid and far from other cluster centroids. (The clustering parameters are e.g., the centroids in K-means or the first two moments in a Gaussian mixture model.) However, this training objective is trivially minimized by the encoder network mapping all data points to the same point in feature space, where this point is equal to one of the cluster centroids. Techniques for avoiding collapse we refer to as partition support. Several partition support methods have been proposed, but they mostly require data augmentation (DA) \cite{zbontar2021barlow, cai2021large}, and those that do not are often ad hoc and lack a rigorous technical foundation \cite{caron2020unsupervised, deshmukh2021representation}. Additionally, they take the soft assignments as a measure of partition collapse and propose to regularize the soft assignments to make them more uniform. This paper focuses on deep clustering without DA, which is an advantage because relying on DA limits a method to domains in which sufficient prior knowledge to perform class-preserving augmentations is available. Additionally, we argue that soft assignments are not an accurate measure of collapse, and that we should instead focus on hard assignments.

We propose a DA-free partition support by regularizing hard assignments. Specifically, we consider the problem of how to optimally assign each data point in a batch to the most appropriate cluster. We express this problem probabilistically in a Bayesian framework, where the regularizing element is captured by a prior across clusters. In the case of equally-sized clusters, this can be uniform. We then use this expression to derive a precise optimization objective, which we also show to be equivalent, up to a small error term, to the objective of maximizing the mutual information of the cluster assignments and the data index. This objective itself is too slow to solve exactly, but we devise a greedy approximation algorithm that can be implemented straightforwardly and results in an intuitive method for fully online clustering, which we term combination assignment. This method outperforms existing online DA-free clustering methods on four popular image clustering datasets. We also analyze the underlying representations and show them to be of high quality. Finally, we analyze the role of hard vs.\ soft cluster assignments in our partition support method, and in previous methods, and make the case that regularizing hard assignments is a more effective approach. Note that, although existing methods can easily convert soft assignments to hard assignments, this is very different from regularizing the hard assignments, as we propose. While the relation between hard and soft clustering has been studied before \cite{bora2014comparative,kearns1998information}, its study in the context of regularizing online deep clustering collapse is new. 

Our main contributions are briefly summarized below.
\begin{itemize}
    \item We articulate a clear Bayesian framework of the problem of hard assignments in online deep clustering models, which avoids the unrealistic assumption of uniformity in each batch made by most existing methods.
    \item We use this framework to derive an optimization objective and prove that it is approximately equivalent to maximizing the mutual information of the cluster assignments and the data index.
    \item We show empirically that the resulting method significantly outperforms existing partition support methods in avoiding partition collapse, improving clustering accuracy and leading to more informative representations.
    \item We conduct further analysis of the performance of different partition support methods, which justifies our choice to focus on hard assignments.
\end{itemize}

The rest of this paper is organized as follows. Section \ref{sec:related-work} gives an overview of related work, while Section \ref{sec:method} lays the theoretical foundations of our method, describes our greedy algorithm for optimizing the resulting objective, and proves the equivalence to mutual information maximization. Section \ref{sec:experimental-eval} reports our empirical results and analysis, and finally Section \ref{sec:conclusion} summarizes our findings.

\section{Related Work} \label{sec:related-work}

A key component of online deep clustering methods is how they avoid the collapsed solution, where (almost) every data point is placed in the same cluster. Methods designed for contrastive learning, and those clustering models that employ it as a part of the training procedure, are more resistant to collapse, because the negative pairs are encouraged to be represented differently. This can mean that they are to be placed in different clusters \cite{huang2021deep}, or just that the encodings should be far apart \cite{zhang2021supporting,cai2021large}. Either approach helps resist all points having the same representation. Even without negative pairs, data augmentation is essential for some methods to avoid collapse. \citet{grill2020bootstrap} perform representation learning using only positive pairs by training an ``online'' network to predict the output of a ``target'' network, which is a slow-moving average of the online network. In \cite{zbontar2021barlow}, data-augmented pairs are used to reduce redundancy by minimizing off-diagonals in the cross-correlation matrix.

Among online clustering models, several partition support methods have been proposed. The solution in \cite{kulshreshtha2018online} is simply to freeze the encoder during clustering, but this requires pretraining it on a separate task offline, so does not work in the fully online setting. A different approach is taken by \citet{gao2020deep}, who use an online autoencoder-based (AE) clustering model, where the reconstruction loss helps to avoid partition collapse, but this limits the method to simple datasets, because the AE's reconstruction loss requires that different images from the same class have significant pixel overlap. In \cite{zhan2020online}, the loss function is continually reweighted to encourage the assignment to smaller clusters. Additionally, when clusters decrease below a certain threshold, they are deleted, and the largest cluster is split in two using K-means, which also means the method does not work fully online. Data-augmented pairs are used by \citet{cai2021large}, with the same method as \cite{zbontar2021barlow}, treating soft cluster assignments as representations, and using data-augmented pairs. In \cite{zhong2020deep}, the sum of squares of the probability (i.e., soft assignment) of each cluster is minimized, marginalized over each training batch. Decomposing the expectation of the square, we see that this also involves minimizing the variance. However, the authors also use contrastive learning and DA to avoid collapse, so they do not fully rely on sum-of-squares minimization.

A similar idea, employed by \cite{li2020prototypical,hu2017learning,niu2021spice,niu2020gatcluster,van2020scan}, is to maximize the entropy of batch-wise marginal soft assignments. For an input distribution $X$ with corresponding soft assigned labels $Y$, both approximated over a batch, an extra term $-H(Y)$ is added to the loss function. As the entropy of a multinomial is maximized at the uniform distribution, this encourages equally sized clusters. Entropy maximization has the advantage of a solid formal interpretation as part of the maximization of the mutual information between data and cluster assignments: maximizing $H(Y)$ but minimizing $H(Y|X)$ (the latter is minimized explicitly in \cite{hu2017learning} and implicitly via contrastive learning in \cite{li2020prototypical}), implicitly maximizes 
\[
I(X;Y) = H(Y) - H(Y|X)\,.
\]
However, the marginal entropy term tends to only be partially successful at preventing partition collapse, often the entire dataset is still put into only a small number of clusters. As well as being confirmed in our experiments in Section \ref{sec:experimental-eval}, this empirical weakness of entropy maximization for avoiding partition collapse is reported in \cite{hu2017learning}, and better results are found by explicitly constraining the soft cluster assignments using non-linear programming. The work \cite{li2020prototypical} does not rely entirely on entropy maximization, because they employ contrastive learning, which (as explained above) also helps to avoid partition collapse. 

Another approach is to directly impose a constraint on cluster assignments. Based on earlier work \cite{asano2019self}, \cite{caron2020unsupervised,deshmukh2021representation,kumar2021unsupervised} proposed to constrain the soft cluster assignments to be marginally uniform across each training batch. Though effective in preventing partition collapse, the constraint of exact uniformity across each batch is too strict, as noted in \cite{kumar2021unsupervised}. The ground-truth classes will almost certainly violate this constraint. 

\section{Method} \label{sec:method}
\paragraph{Problem Formulation} \label{subsec:problem-formulation}
We want to simultaneously (a) train the encoder and (b) make hard assignments to each batch of points at a time, based on the features extracted by that encoder. If the encoder is $f_{\theta_1}: \mathbb{R}^n \rightarrow \mathbb{R}^m$, our clustering model is $g_{\theta_2}:\mathbb{R}^m \rightarrow \{1, \dots, K\}$, parametrized by $\theta_2$, and our batch size is $N$, then we seek a function of the form $\Gamma: \mathbb{R}^{N \times m}\ \rightarrow$ $\{1, \dots, K\}^{N}$. (The difference between $\Gamma$ and $g_{\theta_2}$ is that the former is used during training to assign an entire batch, while the latter is used at inference time and can assign each point individually.) 
If we have a method for batchwise assignment, then the training objective can be formulated as minimizing some notion of distance of points from the centroids of their assigned clusters, and the encoder and clustering model can be trained as follows:
\begin{equation}
    \argmin_{\theta_1, \theta_2} \sum_{i=1}^N d(f_{\theta_1}(x_i) - \mu_{k_i})\,, \label{eq:online-cluster-obj-euclidean}
\end{equation}
where $x_i$ is the $i$th batch elemet, $d(\cdot)$ is a distance function, e.g., Euclidean, $k_i=\Gamma(f_{\theta_1}(x))_i$ is the 
$i$th assigned cluster label, and $\mu_k$ is the $k$th cluster centroid.

Finding a suitable assignment function $\Gamma$ is non-trivial. It should be more likely to assign points to the clusters whose centroids are closer. However, using only this rule, and assigning every vector to its closest centroid, we allow a collapsed solution where \eqref{eq:online-cluster-obj-euclidean} is minimized by $f_{\theta_1}$ mapping every point to a single centroid: $\forall i, f_{\theta_1}(x_i)=c_k$, for some $k\in \{0,\dots,K-1\}$. The heart of our method is the choice of a suitable $\Gamma$, which avoids the collapsed solution. We refer to it as combination assignment, because its prior is the combination of labels under a multinoulli distribution.

\paragraph{Combination Assignment} \label{subsec:comb-assignment}
Combination assignment is based on a Bayesian formulation of the online clustering problem. Let $\mathcal{D}$ be the data distribution, $X \sim \mathcal{D}$ be a sampled batch of data, $Z$ be the random variable defined by applying the feature extraction to $X$ (i.e., $Z$ is the output of a deep encoder network), and let $Y$ be the assigned cluster labels. Here, we consider the encoder to be fixed, and we are just interested in the best hard assignment of cluster labels, given the extracted features. That is, we want to determine the values of $Y$ with the maximum probability under the a posteriori distribution $p(Y|Z)$. We assume we have some reasonable estimate of the prior distribution over $K$ clusters: $p:\{0,\dots,K-1\} \rightarrow [0,1]$ (note, this estimate could change for each batch).
Then, the prior probability of a batch containing exactly $n_k$ labels for each cluster~$k \in\{ 1, \dots, K\}$~is 
\begin{equation} \label{eq:prior}
    \prod_{k=1}^Kp(k)^{n_k} \frac{N!}{\prod_{k=1}^K n_k!}\,,
\end{equation}
where $N=\sum_{k=1}^K n_k$ is the batch size. It is often suitable to choose a uniform prior over $K$ cluster labels, $p(i) = 1/K$, however, we are free to choose any prior we would like. Simply having an estimate of the prior is a weaker assumption than existing methods, which assume equal numbers of each cluster by design. For example, all methods that use k-means as their backbone are implicitly assuming roughly uniform clusters \cite{satapathy2015emerging}. 

We model each cluster as a multivariate normal distribution in feature space, so that the likelihood is given by
\begin{gather}
p(Z=z_1, \dots, z_N|Y=k_1,\dots,k_N) = \\
\prod_{i=1}^N \frac{\exp(-\tfrac{1}{2}(z_i- \mu_{k_i})\Sigma_{k_i}^{-1}(z_i-\mu_{k_i}))}{\sqrt{(2 \pi)^{d}|\Sigma_{k_i}|}}\,,\label{full-likelihood}
\end{gather}
where $d$ is the dimension of the feature space, and $\mu_k$ and $\Sigma_k$ are the centroid and covariance matrix of the $k$th cluster, respectively. 
We then maximize the posterior corresponding to the prior in \eqref{eq:prior} and the likelihood in \eqref{full-likelihood}, giving the following optimization problem (more details in the appendix):
\begin{align} \label{eq:sum-objective}
    &\argmin_Y \sum_{i=1}^N d(z_i,\mu_i,\Sigma_i) - \log{p(k_i)} + \sum_{k=1}^K\log( n_k!), \\
    \text{where  }& d(z,\mu,\Sigma) =\tfrac{1}{2}(z- \mu)^T\Sigma^{-1}(z-\mu) + \tfrac{1}{2}\log{(2\pi)^d|\Sigma|}\,. \notag
\end{align}
In matrix notation, the objective is 
\begin{gather}
    \argmin_{Q \in \mathcal{B}^{N \times K}} \langle Q, \tilde{Q} \rangle - \mathds{1}_N^TQ\log{P} + \log(\mathds{1}_N^TQ!) \mathds{1}_K \label{eq:matrix-objective} \\
    \text{subject to } Q\mathds{1}_K = \mathds{1}_N\,, \notag
\end{gather}
where $\tilde{Q}_{i,j} = d(z_i, \mu_j, \Sigma_j)$, $\langle \cdot , \cdot \rangle$ denotes the Frobenius inner product, $\mathds{1}_a$ is an $a$-dimensional vector of all ones, $P$ is a $K$-dimensional probability vector specifying the prior, and $\log$ and factorial are applied element-wise to the $K$-dimensional vector $\mathds{1}_N^TQ$. Constraining $Q$ to be Boolean enforces hard assignments, constraining $Q\mathds{1}_K = \mathds{1}_N$ enforces each latent vector to be assigned to exactly one cluster. The encoder is then trained with gradient descent w.r.t. using only the distance from the chosen cluster. The prior and cluster sizes affect its training indirectly.

\paragraph{Solving the Optimization Problem} \label{subsec:greedy-strategy}
It is too slow to solve \eqref{eq:matrix-objective} exactly (see appendix for details), but we can motivate an approximation by considering the simpler problem of assigning the last data point in a batch, given that the rest have already been assigned. That is, we maximize the conditional probability of the $N$th assignment in a batch, conditioned on the $N-1$ previous assignments. This gives the following (details in the appendix).
\begin{gather}
    \argmin_{k_N=1,\dots,K} d(z_N,\mu_{k_N},\Sigma_{k_N}) - \log{p(k_N)}+ \log( n_{k_N}+1)\,, \label{eq:Nth-point-assignment}
\end{gather}
where $n_k$ is the size of cluster $k$ before the $n$th assignment.

To approximately solve \eqref{eq:matrix-objective}, we employ a greedy algorithm that iteratively solves \eqref{eq:Nth-point-assignment} with respect to the most confident assignment possible. That is, at each iteration, select the pair $(i,k)$ (corresponding to $(N,k_N)$ above) for which \eqref{eq:Nth-point-assignment} is smallest, with $i$ ranging over the indices of still-unassigned points, and $k$ ranging over all clusters, and assign the $i$th point to cluster $k$.

\paragraph{Intuition}
To get an intuition on our method, recall that clustering generally should assign points to the closest centroid, but that we should try to resist assigning to a cluster that already has lots of points assigned to it and also take into account the prior. Even if such a cluster has its centroid closest to the $N$th point, it may be better to instead assign to a smaller, further away cluster, or one with a higher prior probability. This suggests choosing the cluster assignment so as to minimize a combination of the distance to the centroid, the prior and some increasing function of cluster size, which is precisely what \eqref{eq:Nth-point-assignment} expresses. The first term says to pick a nearby cluster, the second term to pick a cluster with a high prior probability, and the third to penalize clusters that already have many points assigned. It would not be obvious, a priori, what the increasing function of cluster-size should be exactly, but the derivation of \eqref{eq:Nth-point-assignment} shows that $\log{(n+1)}$ is an appropriate choice. The fact that we take into account the cluster size when assigning points to clusters is an important difference between our method and existing methods, which generally assign to the cluster with the closest centroid (e.g., k-means) or the cluster which assigns the highest probability to the given data point (e.g., GMM clustering). Another important difference is that the centroids are trainable parameters, and can be updated by gradient descent, rather than being set to the empirical mean of the corresponding cluster.

\paragraph{Information-Theoretic Interpretation}
Under a uniform prior, the greedy algorithm that iteratively solves \eqref{eq:Nth-point-assignment} can be interpreted as iteratively making whatever assignment will maximize the mutual information between the batch index $i$ and the cluster labels. First, we show a close equivalence to maximizing the entropy of cluster labels in each batch.

Let $H^{(k)}$ be the marginal hard entropy of cluster labels after a new assignment to cluster $k$:
\begin{gather*}
    H^{(k)} = \frac{x_k + 1}{N+1} \log{\frac{x_k + 1}{N+1}} + \sum_{j=1, j\neq k}^K \frac{x_j}{N+1} \log \frac{x_j}{N+1}\,,
\end{gather*}
and consider the difference $ H^{(k)} - H^{(k')}$ between the entropy after making assignment $k$ vs.\ after making a different assignment $k'$. It can be shown (see appendix) that 

\begin{equation} \label{eq:entropy-assignment-objective}
    H^{(k)} - H^{(k')} \approx  \frac{1}{N+1} \left( \log(x_{k'} + 1) - \log (x_{k} + 1)\right)\,.
\end{equation}
Thus, if we were to make each assignment so as to maximize $\log(\mathcal{L}_k(x)) + \lambda H^{(k)}$,
where $\mathcal{L}(k)$ is the likelihood of the new data point under cluster $k$, and $\lambda$ is some hyperparameter, then, subject to the above approximation, for each $k,k' \in \{1, \dots, K\}$, we would prefer to assign to $k$ iff
\begin{gather}
    \log \mathcal{L}(k) + \lambda H^{(k)} > \log \mathcal{L}(k') + \lambda H^{(k')} \iff \notag \\
    \log \mathcal{L}(k) - \log\mathcal{L}(k') >  \nonumber \\ > \frac{\lambda}{N+1} \left( \log(x_{k'} + 1) - \log (x_{k} + 1)\right)\,. \label{eq:entropy-assignment-objective-1}
\end{gather}
Modelling clusters as multivariate normal distributions (as above), and setting $\lambda = N+1$, \eqref{eq:entropy-assignment-objective-1} becomes equivalent to \eqref{eq:Nth-point-assignment}. (See appendix for full proof.)

Thus, our method closely approximates a maximization of the entropy of cluster labels. There is some similarity to those methods, discussed in Section \ref{sec:related-work}, that use an additional loss term to encourage greater entropy of soft assignments in each batch, but an important difference here is that we are maximizing the entropy of hard assignments. This means that the entropy of cluster labels given batch index is automatically zero. Therefore, using the decomposition $I(X;Y) = H(X) - H(X|Y)$, the mutual information of the batch index $i$ and the cluster labels equals the entropy of cluster labels, and so our method is a close approximation to maximizing this mutual information.

\begin{algorithm}[tb]
    \caption{During training, cluster labels are assigned batchwise, with partition support provided by a uniform prior across clusters. During inference, cluster labels are assigned pointwise, without any explicit partition support.} \label{alg:method}
    \begin{algorithmic}
    \State $f_{\theta_1} \gets$ encoder network; 
    \State $\theta_2 = \mu_1, \dots, \mu_K \gets$ centroids for each of the $K$ clusters; 
    \State $\sigma \gets$ isotropic variance of all clusters; 
    \Function{AssignBatch}{Z}   
        \State $counts \gets$ $K$-dimensional array, initially all $0$s; 
        \State $isAssigned \gets$ $N$-dimensional Boolean array, initially all False; 
        \State $D \gets$ $N \times K$ matrix, where $D_{ij} = ||Z_{i} - \mu_{j} ||^2$; 
        \For{r=1,\dots,N}
            \State $\tilde{D} \gets$ $N \times K$ matrix, where $\tilde{D}_{ij} = D_{ij} + 2\sigma \log(counts[j] + 1)$; 
            \State $(a,k) \gets$ argmin $\{\tilde{D}_{ij} : \neg isAssigned[i]\}$; 
            \State $counts[k] \gets counts[k] + 1$; 
            \State $isAssigned[a] \gets True$; 
            \State $loss \gets loss + D_{a,k}$
        \EndFor 
        \State \Return $loss$
    \EndFunction 
    \Function{TrainOnBatch}{X} \funclabel{func:train-method}
        \State $Z \gets f_{\theta_1}(X)$, encodings for a batch of $N$ data points; 
        \State $loss \gets AssignBatch(Z)$;
        \State take a train step on $loss$ with respect to $\theta_1, \theta_2$
    \EndFunction 
    \Function{PredictSingleDataPoint}{x} \funclabel{func:inference-method}
        \State $z \gets f_{\theta_1}(x)$ encodings for a batch of $N$ data points; 
        \State $assignment = \argmin_{j=1,\dots,K} ||z-\mu_j||^2$; 
        \State \Return assignment
    \EndFunction
    \end{algorithmic}
\end{algorithm}

\paragraph{Training Procedure}
Our model comprises an encoder network $f_{\theta_1}$ and cluster centroids $\theta_2 = \{\mu_1, \dots, \mu_K\}$. To train on an input batch, we first encode the raw data using $f_{\theta_1}$, then we employ the combination assignment method of Section \ref{subsec:greedy-strategy}, and use these assignments to minimize \eqref{eq:sum-objective} with respect to both $\theta_1$ and $\theta_2$. As the second two terms in \eqref{eq:sum-objective} have no gradient, the updates are made only with respect to the first term, so similarly to \eqref{eq:online-cluster-obj-euclidean}. At inference time, we do not perform combination assignment. Instead, we simply assign each point to the cluster with the nearest centroid, so the model can assign points individually, and is not restricted to operating batch-wise. The full training and inference methods are described by the functions 
TrainOnBatch and PredictBatch, respectively, in Algorithm~\ref{alg:method}.

\begin{table*}
\centering
\resizebox{0.79\textwidth}{!}{
\begin{tabular}{llllllll}
\toprule
         &     &       CA (ours) &           SK &           ENT &           SS &          CKM &       no reg \\
\midrule
CIFAR 10 & Acc &     \textbf{21.7 (1.50)} &  16.7 (0.40) &   18.6 (1.52) &  11.8 (1.92) &  15.2 (0.91) &  10.0 - \\
         & NMI &    \textbf{10.5 (1.07)} &  03.8 (0.76) &   08.7 (2.94) &  01.1 (1.29) &  02.8 (0.71) &  00.0 - \\
         & ARI &      \textbf{5.4 (0.59)} &  02.7 (0.68) &   05.7 (1.97) &  00.3 (0.42) &  01.4 (0.38) &  00.0 - \\
         & KL* &      \textbf{0.0 (0.01)} &  02.5 (0.33) &   01.8 (0.21) &  00.9 (1.36) &  01.1 (0.24) &  00.0 - \\
CIFAR 100 & Acc &      \textbf{6.9 (0.17)} &  02.6 (0.24) &   02.4 (0.19) &  01.2 (0.25) &  02.8 (0.34) &  01.0 - \\
         & NMI &    \textbf{14.2 (0.61)} &  05.3 (0.46) &   06.3 (0.82) &  00.6 (1.18) &  04.6 (0.89) &  00.0 - \\
         & ARI &      \textbf{1.7 (0.10)} &  00.3 (0.04) &   00.5 (0.11) &  00.0 (0.04) &  00.3 (0.12) &  00.0 - \\
         & KL* &      \textbf{0.5 (0.11)} &  03.5 (0.80) &   01.6 (0.22) &  00.2 (0.19) &  02.8 (0.49) &  00.0 - \\
FashionMNIST & Acc &     \textbf{59.3 (4.16)} &  25.1 (3.13) &   25.5 (6.16) &  10.0 (0.04) &  22.0 (3.83) &  10.0 - \\
         & NMI &     \textbf{55.3 (2.95)} &  17.7 (2.33) &  20.9 (11.32) &  00.0 (0.04) &  15.8 (5.28) &  00.0 - \\
         & ARI &     \textbf{42.5 (4.12)} &  09.2 (1.42) &   10.6 (6.01) &  00.0 - &  07.3 (2.84) &  00.0 - \\
         & KL* &      \textbf{0.0 (0.02)} &  02.6 (0.35) &   01.9 (0.24) &  00.0 (0.01) &  01.3 (0.31) &  00.0 - \\
STL & Acc &     \textbf{24.2 (2.56)} &  18.5 (1.08) &   10.9 (1.97) &  10.1 (0.22) &  14.1 (0.78) &  10.0 - \\
         & NMI &     \textbf{13.7 (1.32)} &  06.7 (1.14) &   01.3 (2.86) &  00.0 (0.09) &  02.5 (0.58) &  00.0 - \\
         & ARI &      \textbf{7.3 (1.11)} &  03.3 (0.88) &   00.2 (0.36) &  00.0 - &  01.1 (0.39) &  00.0 - \\
         & KL* &      \textbf{0.1 (0.10)} &  00.7 (0.41) &   00.1 (0.18) &  00.1 (0.18) &  01.2 (0.21) &  00.0 - \\
RealDisp & Acc &     \textbf{57.0 (3.02) }&  23.5 (3.09) &   18.8 (1.98) &  10.7 (1.88) &  17.2 (1.33) &  10.0 - \\
         & NMI &  \textbf{55.0 (17.11)} &  31.0 (2.93) &   29.7 (4.26) &  00.5 (0.50) &  25.8 (1.75) &  00.0 - \\
         & ARI &     \textbf{48.1 (4.49)} &  11.6 (3.02) &   06.6 (1.99) &  00.2 (0.26) &  06.4 (1.91) &  00.0 - \\
         & KL* &      \textbf{0.1 (0.06)} &  00.7 (0.13) &   01.7 (0.61) &  02.0 (0.91) &  01.6 (0.13) &  02.3 - \\
\bottomrule
\end{tabular}

}
\caption{Effect, on cluster size and clustering performance, of our method compared to two existing partition-support methods. 
        All figures are the mean of 5 runs, with std dev in parentheses. Best results in bold, std dev of zero after rounding, is written `-'.}\label{tab:main-results}
\end{table*}
\FloatBarrier

\section{Experimental Evaluation} \label{sec:experimental-eval}

\paragraph{Datasets and Metrics}
We report results on CIFAR 10 (C10), CIFAR 100 (C100), FashionMNIST (FMNIST), and STL, with image sizes 32, 32, 28, and 96, respectively, and the human activity recognition (HAR) dataset RealDisp, of 17 subjects performing 33 different activities wearing accelerometers. We use the standard clustering metrics of accuracy (ACC), normalized mutual information (NMI), and adjusted Rand index (ARI), defined as, e.g., in \cite{sheng2020unsupervised}. We also report the KL-divergence from the ground truth of the model's empirical distribution over clusters, denoted KL$^*$. For a collapsed model, which assigns most points to a few clusters, this will be high.

\subsection{Clustering Accuracy and Degree of Collapse} \label{subsec:main-results}
Table \ref{tab:main-results} compares our method with four existing methods: sum of squares minimization, denoted ``SS'' \cite{zhong2020deep}, the Sinkhorn-Knopp algorithm for optimal transport, denoted ``SK'' \cite{caron2020unsupervised,kumar2021unsupervised}, and marginal entropy maximization \cite{li2020prototypical}, denoted ``Ent'' and concrete k-means, denoted ``CKM'' \cite{gao2020deep}. Each is described in Section \ref{sec:related-work}. To make further explicit the phenomenon of partition collapse, we also include a model without any partition support. Our method significantly outperforms others on all datasets and metrics. 

Observe that the unregularized model exhibits total collapse in all experiments, placing all points in the same cluster and consequently achieving a cluster performance no better than random guessing. The performance of SS is not much better. By making a slight change to the author's original method, we could actually significantly improve its results (see appendix), but it was still unreliable and less accurate than our method.  The other two existing partition support methods do a reasonable job of avoiding partition collapse. However, entropy maximization occasionally also reaches the state with all points in the same cluster (this is consistent with previous literature, e.g., \cite{hu2017learning}). The Sinkhorn-Knopp method is more reliable, but by far the most uniform cluster sizes are produced by our method. Note that we do not employ our assignment algorithm at inference time,  and instead assign each point to the cluster with the nearest centroid. This shows that our cluster centroids are well-distributed around the data manifold, each capturing a sizeable subset of the data even when the explicit support is removed. Together, these figures show that (a) partition support is necessary to learn anything meaningful, (b) our method of combination assignment is better at avoiding partition collapse than previous methods, and (c) our model has, consequently, better clustering performance.

To better isolate the effect of our proposed partition support method, we do not perform hyperparameter tuning, and use a relatively simple architecture for all datasets. This is the same for all methods being compared. The network consists of two convolutional layers with filter sizes $6$ and $16$,  and ReLU activations, followed by a fully connected layer to a latent space of dimension $128$. Training uses Adam, with learning rate 1e-3, $\beta_1=0.9$, $\beta_2 = 0.99$, and batch size 256. We follow previous works in setting $K$, the number of clusters, to the number of ground-truth classes. We set $\Sigma=\sigma I$ with $\sigma=1e2$. We observe very similar results for a wide range of values for $\Sigma$, including a setting where we continually estimate from the empirical distributions. 

Although some existing clustering methods produce higher accuracy than those of Table \ref{tab:main-results}, this is not a valid comparison, because (as discussed above) these higher-scoring methods (a) use data-augmentation and (b) even more significantly, address offline clustering, which is much easier than, and not comparable to, the online cluster we address.

 \subsection{Imbalanced Data} \label{subsec:imbalanced-data}
\begin{table}
\resizebox{\columnwidth}{!}{
\begin{tabular}{lllll}
\toprule
 &  & imb 1 & imb 2 & imb 3 \\
\midrule
\multirow[t]{4}{*}{CIFAR 10} & acc & 22.2 (1.22) & 22.8 (1.46) & 24.6 (1.26) \\
 & nmi & 10.5 (1.14) & 11.3 (0.73) & 10.5 (0.50) \\
 & ari & 5.9 (0.92) & 6.3 (0.55) & 6.2 (3.22) \\
 & KL* & 0.1 (0.07) & 0.1 (0.04) & 0.2 (0.11) \\
\cline{1-5}
\multirow[t]{4}{*}{CIFAR 100} & acc & 6.6 (0.31) & 6.7 (0.62) & 9.0 (0.33) \\
 & nmi & 13.3 (0.62) & 13.7 (0.99) & 15.4 (0.30) \\
 & ari & 1.6 (0.19) & 1.9 (0.36) & 2.7 (0.27) \\
 & KL* & 0.8 (0.10) & 0.7 (0.10) & 0.6 (0.09) \\
\cline{1-5}
\multirow[t]{4}{*}{FMNIST} & acc & 54.6 (6.34) & 52.1 (3.39) & 53.8 (2.95) \\
 & nmi & 52.9 (4.41) & 50.6 (1.50) & 50.0 (2.35) \\
 & ari & 37.7 (5.56) & 36.7 (1.84) & 38.5 (3.08) \\
 & KL* & 0.2 (0.23) & 0.1 (0.05) & 0.2 (0.17) \\
\cline{1-5}
\multirow[t]{4}{*}{STL} & acc & 23.2 (2.07) & 24.6 (1.41) & 25.5 (3.13) \\
 & nmi & 13.3 (1.69) & 14.5 (1.15) & 12.6 (2.44) \\
 & ari & 6.8 (1.13) & 7.8 (0.81) & 6.8 (1.90) \\
 & KL* & 0.3 (0.20) & 0.2 (0.19) & 0.1 (0.03) \\
\cline{1-5}
\multirow[t]{4}{*}{RealDisp} & acc & 57.5 (1.45) & 46.7 (6.75) & 39.2 (6.20) \\
 & nmi & 75.1 (1.01) & 67.9 (5.65) & 62.9 (5.77) \\
 & ari & 49.4 (0.90) & 38.0 (6.66) & 30.0 (6.07) \\
 & KL* & 0.1 (0.07) & 0.0 (0.01) & 0.3 (0.03) \\
\cline{1-5}
\bottomrule
\end{tabular}
}
\caption{Performance on imbalanced data, imb. 1 removes $10\%$ of one class and $20\%$ of another, imb. 2 removes $0\%, 5\%, 10\%, \dots$ of each of the 10 C10 classes, imb.3 is the same but in increments of $10\%$ (scaled proportionally to class size for other datasets). Note, a random guess will score higher with increasing imbalance. } \label{tab:imbalanced-results}
\end{table}

As described in Section \ref{sec:method}, our method can use any prior distribution over clusters, not just a uniform distribution (i.e., equal cluster sizes) as most existing methods assume. Here, we conduct an empirical investigation of performance on imbalanced data. We assume that, in this case, we have an idea of the relative frequency of each class/cluster and that this determines our prior. To explore varying levels of class-imbalance, we begin with the balanced datasets from Section \ref{subsec:main-results}, and progressively remove parts of some classes, giving three increasing levels of imbalance. As shown in Table \ref{tab:imbalanced-results}, our method does not degrade in performance over these settings, showing robustness to varying class distributions.

\begin{table*}
\resizebox{0.68\textwidth}{!}{
\begin{tabular}{lllllll}
\toprule
         &      &    CA (ours) &           SK &          ENT &            SS &          CKM \\
\midrule
C10 & linear &  \textit{35.5 (0.73)} &  30.0 (0.66) &  35.4 (1.53) &   26.0 (2.39) & \textbf{ 40.2 (0.30)} \\
         &  KNN &  30.3 (0.87) &  27.1 (1.02) &  \textit{33.3 (2.29)} &   22.8 (1.74) &  \textbf{38.3 (0.61)} \\
C100 & linear &  \textbf{17.1 (0.08)} &   8.9 (1.74) &   6.1 (0.77) &    7.9 (1.03) &  \textit{16.9 (0.48)} \\
         &  KNN &  \textit{12.8 (0.39)} &   8.4 (1.88) &   5.3 (0.86) &    7.7 (1.04) &  \textbf{13.7 (0.73)} \\
FM & linear &  79.3 (1.83) &  75.8 (1.39) &  \textit{80.1 (1.36)} &   62.8 (2.55) &  \textbf{81.9 (0.53)} \\
         &  KNN &  77.0 (1.76) &  \textit{80.5 (0.58)} &  80.4 (1.09) &   78.5 (1.53) &  \textbf{81.3 (0.80)} \\
STL & linear &  \textit{36.0 (2.03)} &  32.4 (0.76) &  33.6 (0.80) &   33.0 (1.57) &  \textbf{40.4 (1.49)} \\
         &  KNN &  \textbf{36.8 (1.85)} &  27.3 (3.33) &  27.5 (0.70) &   29.0 (1.49) &  \textit{35.6 (1.74)} \\
RD & linear &  \textbf{95.9 (0.58)} &  79.7 (5.21) &  38.5 (6.93) &   31.1 (2.44) &  \textit{83.4 (1.77)} \\
         &  KNN &  \textbf{98.8 (0.23)} &  \textit{97.1 (1.87)} &  93.4 (1.03) &  40.8 (28.58) &  88.7 (2.24) \\
\bottomrule
\end{tabular}

}
\centering
\caption{Quality of the learned representations, as assessed by linear and KNN probes. Our method performs comparably to CKM on the vision datasets, and better on RealDISP, fitting with the case made in \citet{mahon2022har} against autoencoders in HAR clustering.} \label{tab:probe-results}
\end{table*}

\subsection{Hard vs.\ Soft Assignment Regularization}
\begin{figure}[h]
    \centering
    \includegraphics[width=\columnwidth]{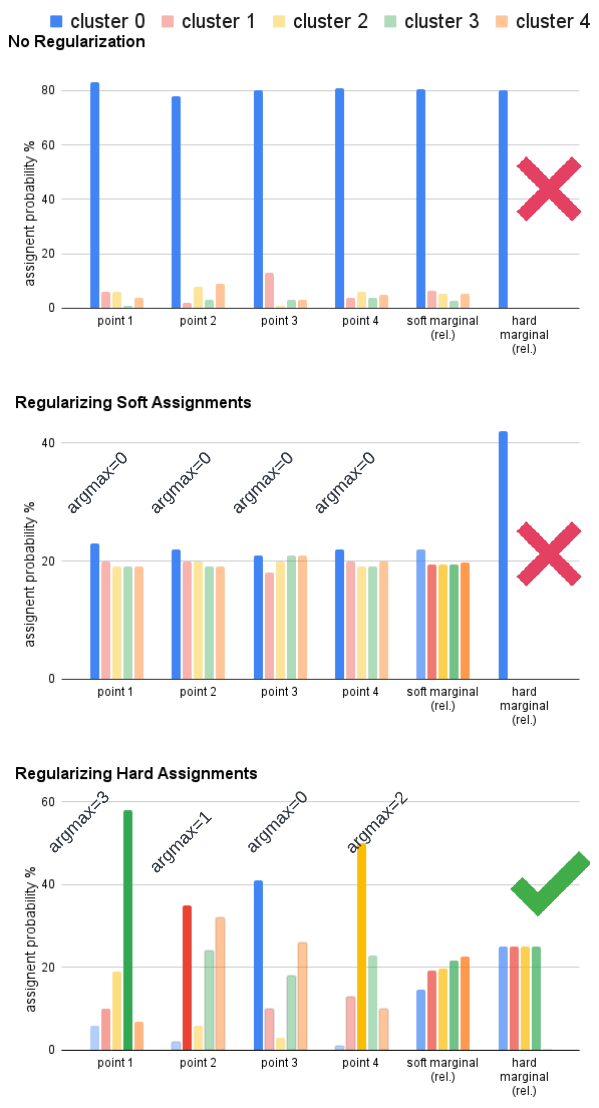}
    
    \caption{
   Top: with no regularization, the model collapses. 
    Middle: soft regularization encourages uniform soft assignments but the argmax is constant. 
    Bottom: hard regularization causes the argmax to change, and avoids collapse.}
    \label{fig:hard-soft-comparison}
\vspace{-1.5ex}
\end{figure}

A key element of our method is the regularization of hard assignments, whereas previous methods regularize soft cluster assignments. Ours is a fundamentally different form of regularization, and one which we argue is better able to prevent collapse. For example, assume there are only three clusters and a batch size of four, and consider the following two matrices of assignment probabilities
\[
D_1 = \begin{bmatrix}
.98 & .01 & .01 \\
.98 & .01 & .01 \\
.49 & .50 & .01 \\
.49 & .01 & .50 
\end{bmatrix}
D_2 = \begin{bmatrix}
.34 & .33 & .33 \\
.34 & .33 & .33 \\
.34 & .33 & .33 \\
.34 & .33 & .33 
\end{bmatrix}  \,.
\]
The hard and soft entropy (i.e., the entropy of marginal hard and soft assignments) are $1.5$ and $1.1$ for $D_1$, and $0$ and $1.58$ for $D_2$. That means $D_2$ has higher soft entropy but a much lower hard entropy. Indeed, despite having nearly maximum soft entropy, $D_2$ is collapsed with zero hard entropy. A similar scenario is shown for an idealized batch in Figure \ref{fig:hard-soft-comparison}, again revealing that near-uniform soft assignments is a different objective to, and does not guarantee, avoiding collapse.

To investigate whether such differences between hard vs.\ soft regularization manifest in practice, we empirically compare the KL$^*$ for the hard and soft assignments. The results are shown in Table \ref{tab:entropy-comparison}.
The most striking difference between hard and soft entropy is in SS. There, the soft KL$^*$ is often close to the maximum value (equal to the logarithm of the number of clusters), but the hard entropy is consistently close to zero. This shows that this form of regularization produces batch assignment probabilities similar to matrix $D_2$ above, where the probabilities for each data point are squeezed close to one another, without much change in the order of highest to lowest, in particular the argmax.

\begin{table}

\resizebox{\columnwidth}{!}{
\begin{tabular}{@{\,}l*{5}{l}l@{\,}}
\cline{1-7}
    &              &  CA &       SK &      Ent &       SS &     \!\!CKM \\
\cline{1-7}
\multirow{2}{*}{CIFAR 10}
    & hard KL$^*$ &       \textbf{0.04} &     0.56     &     1.05     &     1.66 & 1.13 \\
    & soft KL$^*$ &       0.04 &     \textbf{0.00} &     0.85     &     0.51 & 0.03\\
\cline{1-7}
\multirow{2}{*}{CIFAR 100} 
    & hard KL$^*$ &       \textbf{0.83} &     2.17     &     3.48     &     4.47 & 2.78 \\
    & soft KL$^*$ &       0.13 &     \textbf{0.01} &    1.42     &     0.09 & 0.02\\
\cline{1-7}
\multirow{2}{*}{FMNIST} 
    & hard KL$^*$ &       \textbf{0.04} &     0.47     &     0.99     &     2.30 & 1.32 \\
    & soft KL$^*$ &       0.04 &     \textbf{0.00} &    0.69     &     0.17  & 0.02 \\
\cline{1-7}
\multirow{2}{*}{STL} 
    & hard KL$^*$ &       \textbf{0.09} &     2.31 &     2.25     &     2.25 & 1.17 \\
    & soft KL$^*$ &       0.08 &     \textbf{0.00} &     0.92     &     0.18 & 0.02 \\
\cline{1-7}
\multirow{2}{*}{RealDisp} 
    & hard KL$^*$ &       \textbf{0.05} &     0.66 &     1.69     &   2.05   & 1.65 \\
    & soft KL$^*$ &       0.00 &     \textbf{0.00} &     0.08     &     0.00 & 0.02 \\
\cline{1-7}
\end{tabular}
}
       \caption{Comparison of KL$^*$ for hard and soft assignments. Previous methods regularize soft assignments, but this does not transfer well to hard assignments. Ours (CA) is the only method that closely approximates the true distribution for hard assignments. Lower is better, best results in bold.}
    \label{tab:entropy-comparison}
\vspace{-1.5ex}
\end{table}

A similar discrepancy is found in SK, which produces near-perfect uniformity in the soft assignments, and with maximum entropy (up to rounding) on each dataset. This is because it is a (close approximation to a) hard constraint problem. However,  SK's hard assignments still show significant variability, and markedly lower entropy than the soft assignments, suggesting it also produces batch assignment probabilities similar to $D_2$ above. Our method, on the other hand, explicitly forces the argmax to be more evenly distributed during training and, as Table \ref{tab:entropy-comparison} shows, this transfers to the testing setting as well. (Recall that, during testing, we simply assign each point to the cluster with the nearest centroid.) Our hard entropy is only slightly lower than our soft entropy, and is consistently higher than that of the other three methods. This supports our argument that the mean soft assignments do not contain sufficient information to determine if the clustering model is learning a meaningful partition, and regularizing this quantity is not an optimal way to prevent collapse. The distribution of the argmax is also important and is not captured by mean soft assignment.

Our analysis here of hard vs.\ soft cluster assignments does not contradict \citet{caron2020unsupervised}. They report better results using soft assignments as training labels, while we show the superiority of \emph{regularizing}, i.e., encouraging equal numbers of, hard assignments. The two are different contexts of hard and soft labels. We also explored training SK (the method used by \citeauthor{caron2020unsupervised}) using soft assignments as targets, but the results were slightly worse than using hard targets. 

\subsection{Quality of Learned Representations} \label{subsec:learned-representations}
Although ours is primarily a clustering method, we also examine the quality of the learned representations, using linear-probe protocol \cite{zhang2017split}, which trains a linear model and KNN model ($k=10$) to predict the class labels from encoded feature vectors. The results are shown in Table \ref{tab:probe-results}. It is expected that CKM performs well in this setting, because it uses an autoencoder, which is a widely used method for representation learning. Still, however, our method performs on par with CKM, and outperforms the other methods. Note also CKM's poor performance on RealDisp, suggesting that the effectiveness of a decoder may be specific to simple 
vision~datasets.

\section{Conclusion} \label{sec:conclusion}
This paper proposed a data-augmentation-free method to prevent collapse in online deep clustering. We frame probabilistically the problem of deciding which clusters to assign a batch of data points to, given the cluster centroids and features of the data points, and hence derive an intuitve optimization objective for making hard cluster assignments. We then described an algorithm to approximately solve this optimization problem and demonstrated empirically on four datasets that this method outperforms existing methods, both in preventing collapse and in the resulting clustering performance. Finally, we analyzed how the cluster distribution is affected by our partition support method and previous comparable methods. The analysis suggests that regularizing the soft assignments, as is done by existing works, is not sufficient to prevent collapse, and that a better approach is to regularize the hard assignments, as is done by our method.

\section*{Acknowledgments}
This work was supported by the
AXA Research Fund.
{\small
\bibliography{bibliography}
}

\appendix

\onecolumn
\section{Full Proofs and Derivations}
\subsection{Derivation of Main Optimization Objective}
The prior probability of a batch containing exactly $n_k$ labels for each cluster~$k \in\{ 1, \dots, K\}$ is 
\begin{equation*}
    \prod_{k=1}^Kp(k)^{n_k} \frac{N!}{\prod_{k=1}^K n_k!}\,,
\end{equation*}
where $p(\cdot)$ is the prior. As the distribution within each cluster is modelled as multivariate normal, we can write the likelihood as 
\begin{gather*}
p(Z=z_1, \dots, z_N|Y=k_1,\dots,k_N) = \\ \notag
\prod_{i=1}^N \frac{\exp(-\tfrac{1}{2}(z_i- \mu_{k_i})\Sigma_{k_i}^{-1}(z_i-\mu_{k_i}))}{\sqrt{(2 \pi)^{d}|\Sigma_{k_i}|}}\,,\label{full-likelihood-appendix}
\end{gather*}
where $d$ is the dimension of the feature space, and $\mu_k$ and $\Sigma_k$ are the centroid and covariance matrix of the $k$th cluster, respectively. 

Then the full a posteriori is
\begin{align*}
    p(Y|Z) \propto P(Y)P(Z|Y) =&
    \prod_{k=1}^Kp(k)^{n_k} \frac{N!}{\prod_{k=1}^K n_k!}\prod_{i=1}^N \frac{\exp(-\tfrac{1}{2}(z_i- \mu_{k_i})\Sigma_{k_i}^{-1}(z_i-\mu_{k_i}))}{\sqrt{(2 \pi)^{d}|\Sigma_{k_i}|}}\, \\[8pt] 
    \propto & \prod_{k=1}^Kp(k)^{n_k} n_k!\prod_{i=1}^N \frac{\exp(-\tfrac{1}{2}(z_i- \mu_{k_i})\Sigma_{k_i}^{-1}(z_i-\mu_{k_i}))}{\sqrt{(2 \pi)^{d}|\Sigma_{k_i}|}}\,,
\end{align*}
where we drop the constants that are independent of $Y$. Letting $d(z,\mu,\Sigma) =\tfrac{1}{2}(z- \mu)^T\Sigma^{-1}(z-\mu) + \tfrac{1}{2}\log{|\Sigma|} + \tfrac{d}{2}\log{2\pi}$, then, we obtain an optimization objective by minimizing the corresponding negative log-likelihood as follows
\begin{align*}
\argmax_Y p(Y|Z) &\propto \argmax_Y p(Y)P(Z|Y) = \notag \\
 &=   \argmax_Y \log\left(\prod_{k=1}^Kp(k)^{n_k}\right) +  \log\left(\prod_{i=1}^N  \exp(-d(z_i,\mu_{k_i},\Sigma_{k_i})\right)\log(\prod_{k=1}^K n_k!) = \notag \\
 &=   \argmax_Y \sum_{k=1}^Kn_k\log{p(k)}+ \sum_{i=1}^N -d(z_i,\mu_{k_i},\Sigma_{k_i}) - \sum_{k=1}^K\log( n_k!) = \notag \\
 &=   \argmax_Y \sum_{i=1}^N\log{p(k_i)}+ \sum_{i=1}^N -d(z_i,\mu_{k_i},\Sigma_{k_i}) - \sum_{k=1}^K\log( n_k!) = \notag \\
 &=   \argmin_Y \sum_{i=1}^N (d(z_i,\mu_{k_i},\Sigma_{k_i}) - \log{p(k_i)}) +  \sum_{k=1}^K\log( n_k!)\,. 
\end{align*}

\subsection{Computational Complexity}
The main optimization objective is too slow to solve exactly. A common solution would involve interpreting the problem as the rectangular assignment problem, where clusters are workers and data points are jobs. Then take the standard representation of the assignment problem as a flow network. Instead of adding one edge from the source vertex for each worker, add $m$ parallel edges for each worker. For $k \in \{0, \dots, m-1\}$ the , $k$th edge for a worker has capacity 1 and cost $\log{k!} - \log{(k-1)!} = \log{k}$. However, using standard solutions to the assignment problem would then result in complexity cubic in $m$, which is the batch size. This would be prohibitively slow for all but very small batch sizes. 

If $C_{Enc}$ is the cost of a forward and backward pass of the encoder, $N$ is the batch size and $M$ is the latent dimension, then the complexity of our method is $C_{Enc} + \theta(N\log{NM} + NM + N^2)$, because for each of the $NM$ elements in the matrix specifying the cost of assigning each batch element to each cluster, it needs to extract the max and then update the costs with the cluster counts. In comparison, the complexity for SK is $C_{Enc} + \theta{nNM}$, where $n$ is the number of iterations performed in the Sinkhorn Knopp algorithm, CKM is $C_{Enc} + \theta(NM) + C_{Dec}$.

\subsection{Derivation of Greedy Approximation}
We want to maximize the conditional probability of the $N$th assignment in a batch, conditioned on the $N-1$ previous assignments:
\begin{gather}
    \argmax_{k_N=1,\dots,K} p(y_N=k|y_1 = k_1, \dots, y_{N-1}=k_{N-1}; Z) = \notag \\[8pt]
    \argmax_{k_N=1,\dots,K}\frac{p(y_1 = k_1, \dots, y_{N}=k_{N} | Z)}{p(y_1 = k_1, \dots, y_{N-1}=k_{N-1} | Z)} = \notag \\[15pt]
    \argmax_{k_N=1,\dots,K}\log p(y_1 = k_1, \dots, y_{N}=k_{N}| Z) - \notag \\
        - \log p(y_1 = k_1, \dots, y_{N-1}=k_{N-1} | Z) = \notag \\[12pt]
    \argmax_{k_N=1,\dots,K}-\sum_{i=1}^N (d(z_i,\mu_{k_i},\Sigma_{k_i}) + \log{p(k_i)}) + \sum_{k=1}^K\log( n_k'!) + \notag \\
        +\sum_{i=1}^{N-1} (d(z_i,\mu_{k_i},\Sigma_{k_i}) + \log{p(k_i)}) - \sum_{k=1}^K\log( n_k!) = \notag \\[8pt]
    \argmin_{k_N=1,\dots,K} \sum_{i=1}^N (d(z_i,\mu_{k_i},\Sigma_{k_i}) - \log{p(k_i)}) - \sum_{i=1}^{N-1} (d(z_i,\mu_{k_i},\Sigma_{k_i}) - \log{p(k_i)}) + \notag \\
        + (\sum_{k=1}^K\log( n_k'!) - \sum_{k=1}^K\log( n_k!) ) \notag = \\[8pt]
    \argmin_{k_N=1,\dots,K} d(z_N,\mu_{k_N},\Sigma_{k_N}) - \log{p(k_N)} + (\sum_{k=1}^K\log(n_k!) - \sum_{k=1}^K\log( n_k!')) \label{eq:Nth-point-assignment-unsimplified}\,,
\end{gather}
where $n_k$ is the number of points assigned to cluster $k$ before the $N$th assignment, and $n_k'$ is the number assigned to the $k$th cluster after all assignments have been made. This means that
\[
n_{k}' = 
\begin{cases}
    n_{k}+1  & k = k_N \\
    n_{k}  & \text{otherwise}\,. 
\end{cases}
\]
Thus, \eqref{eq:Nth-point-assignment-unsimplified} becomes 
\begin{gather}
    \argmin_{k_N=1,\dots,K} d(z_N,\mu_{k_N},\Sigma_{k_N}) - \log{p(k_N)}+ \log( n_{k_N}+1!) - \log( n_{k_N}!) = \notag \\
    \argmin_{k_N=1,\dots,K} d(z_N,\mu_{k_N},\Sigma_{k_N}) - \log{p(k_N)}+ \log( n_{k_N}+1)\,. 
\end{gather}

\subsection{Proof of Equivalence to Mutual Information Maximization}
We want to show that, in the case of a uniform prior, the greedy algorithm that iteratively solves \eqref{eq:Nth-point-assignment} can be interpreted as (a close approximation to) iteratively making whatever assignment will maximize the mutual information between the batch index $i$ and the cluster labels. First note that, because the proposed model makes hard assignments, the entropy of cluster labels given the batch index is automatically zero, and so recalling that 
\[
I(X;Y) = H(X) - H(X|Y) = H(Y) - H(Y|X)\,,
\]
we see that the mutual information of the batch index $i$ and the cluster labels equals the entropy of cluster labels. Below, we show that the proposed method, up to small approximation error, maximizes the  entropy of cluster labels and, hence, the mutual information of cluster labels and the batch indices in each batch.

\begin{lemma} \label{lemma:ent-diff}
Let $X \in \mathbb{R}^{N \times K}$ be the matrix of already-made assignments in the current batch, and let $H^{(k)}$ be the marginal entropy after the new hard assignment is made to cluster $k$. Then 
\begin{equation}
    H^{(k)} - H^{(k')} \approx  \frac{1}{N+1} \left( \log(x_{k'} + 1) - \log (x_{k'} + 1)\right)\,.
\end{equation}
\end{lemma}

\begin{proof}
Let $X \in \mathbb{R}^{N \times K}$ be the matrix of already-made assignments in the current batch after $N$ points have been assigned, so that $H$, the current marginal entropy of $X$, is given by:
\begin{gather*}
    H = - \sum_{j=1}^K \left(\frac{1}{N} \sum_{i=1}^N x_{ij}\right) \log\left(\frac{1}{N}\sum_{i=1}^Nx_{ij}\right)\,.
\end{gather*}
To simplify notation, let $x_j = \sum_{i=1}^N x_{ij}$. Then $H^{(k)}$, the marginal entropy after the new hard assignment is made to cluster $k$, is given by
\begin{align*}
    H^{(k)} =& - \frac{x_k + 1}{N+1} \log{\frac{x_k + 1}{N+1}} - \sum_{j=1, j\neq k}^K \frac{x_j}{N+1} \log \frac{x_j}{N+1} \\
    =& \frac{-1}{N+1} \left((x_k+1) \log (x_k + 1) + \sum_{j=1, j\neq k}^K x_j (\log x_j -
    \log(N+1))\right) \\
    =& \frac{-1}{N+1} \left((x_k+1) \log (x_k + 1) + \sum_{j=1, j\neq k}^K x_j \log x_j -\sum_{j=1, j\neq k}^K x_j\log(N+1)\right) \\
    =& \frac{-1}{N+1} \left((x_k+1) \log (x_k + 1) + \sum_{j=1, j\neq k}^K x_j \log x_j -
    N\log(N+1)\right) \\
    =& \frac{-1}{N+1} \left((x_k+1) \log (x_k + 1) + \sum_{j=1, j\neq k}^K x_j \log x_j\right) + \frac{N}{N+1}\log(N+1)  \\
\end{align*}
Now, consider the difference $ H^{(k)} - H^{(k')}$ between the entropy after making assignment $k$ vs.\ after making a different assignment $k'$.

\begin{align*}
    =& \frac{-1}{N+1} \left((x_k+1) \log (x_k + 1) + \sum_{j=1, j\neq k}^K x_j \log x_j \right)- \\
    &\frac{-1}{N+1} \left((x_{k'}+1) \log (x_{k'} + 1) + \sum_{j=1, j\neq k'}^K x_j \log x_j \right)  = \\
    =& \frac{-1}{N+1} \left( (x_k+1) \log (x_k + 1) - (x_{k'} + 1)\log (x_{k'} + 1) \right)+ \\
    &\frac{-1}{N+1} \left( \sum_{j=1, j\neq k}^K x_j \log x_j - \sum_{j=1, j\neq k'}^K x_j \log x_j \right)  = \\
    =& \frac{-1}{N+1} \big( (x_k+1) \log (x_k + 1) - (x_{k'} + 1)\log (x_{k'} + 1) \big)+ 
    \left( x_{k'} \log x_{k'} -  x_k \log x_k \right)  = \\
    =& \frac{-1}{N+1} \left(((x_k + 1)\log(x_k + 1) - x_k \log x_k) - ((x_{k'} + 1)\log(x_{k'} + 1) - x_{k'} \log x_{k'})\right) \\
    \approx& \frac{-1}{N+1} \left( (\log(x_k + 1) + \frac{x_k}{x_k + 1}) - (\log (x_{k'} + 1)  + \frac{x_{k'}}{x_{k'} + 1})\right) \\
    =& \frac{-1}{N+1} \left( \log(x_{k} + 1) - \log (x_{k'} + 1) - \frac{x_{k} - x_{k'}}{(x_k + 1)(x_{k'} + 1)}\right) \\
    =& \frac{1}{N+1} \left( \log(x_{k'} + 1) - \log (x_{k} + 1) - \frac{x_{k'} - x_{k}}{(x_k + 1)(x_{k'} + 1)}\right) \\
    =& \frac{1}{N+1} \left( \log(x_{k'} + 1) - \log (x_{k} + 1)\right) - \frac{1}{N+1}\left(\frac{x_{k'} - x_{k}}{(x_k + 1)(x_{k'} + 1)}\right) \\
\end{align*}
where the fourth last line uses the fact that $\log n \approx H_n$ to make the substitution
\[
x_k\log x_k \approx x_k (\log (x_k + 1) - \frac{x_k}{x_k + 1}) \,,
\]
and similarly for $x_{k'}$. Note that the term $\frac{1}{N+1}\frac{x_k - x_{k'}}{(x_k + 1)(x_{k'} + 1)}$ is $0$ in expectation and has absolute value $\leq \frac{N}{(N+1)^2}$. If we drop this small error term, then we get
\begin{equation} 
    H^{(k)} - H^{(k')} \approx  \frac{1}{N+1} \left( \log(x_{k'} + 1) - \log (x_{k} + 1)\right)\,,
\end{equation}
as desired.
\end{proof}

\begin{lemma} \label{lemma:rework-ent-optimization-problem}
Assume that $N$ data points in a batch have already been assigned. Let $\mathcal{L}(k)$ be the posterior probability of the batch of data, under the a $K$-component Gaussian mixture model with isotropic variance $\sigma^2$ (as described in Section \ref{subsec:comb-assignment}), after the $(N+1)$th data point is assigned to cluster $k$. Let $H^k$ be, as above, the entropy of cluster sizes after the $(N+1)$th data point has been assigned to cluster $k$. Then maximizing the objective $\log{\mathcal{L}(k)} + \lambda H^k$ with respect to the $(N+1)$th cluster assignment gives the following optimization problem
\[
\argmin_{k \in \{1,\dots,K\}} d(z_i,\mu_i,\Sigma_i) + \frac{\lambda}{N+1} \log (x_{k} + 1)
\]
\end{lemma}

\begin{proof}
Maximizing $\log{\mathcal{L}(k)} + \lambda H^k$  with respect to the $(N+1)$th cluster assignment means we prefer to assign to cluster $k$ over cluster $k'$ if and only if
\begin{gather}
    \log{\mathcal{L}(k)} + \lambda H^k > \log{\mathcal{L}({k'})} + \lambda H^{k'} \iff \\
    \log{\mathcal{L}(k)} - \log{\mathcal{L}({k'})} > \lambda H^{k'} - \lambda H^{k} \iff \\
    -\log{\mathcal{L}(k)} - (-\log{\mathcal{L}({k'}))} < \lambda H^k - \lambda H^{k'} \iff \\
    \log{\mathcal{L}(k)} - \log{\mathcal{L}({k'})} <  \frac{\lambda}{N+1} \left( \log(x_{k'} + 1) - \log (x_{k} + 1)\right) \,, \label{eq:difference-NLL-ent-objective1}
\end{gather}
where the last line uses lemma \ref{lemma:ent-diff}. Let $z$ be the encoding of the $(N+1)$th point, as in Section \ref{subsec:comb-assignment}. Then
\begin{align*}
-\log{\mathcal{L}(k)} =& -\log{ \left(\frac{\exp(-\tfrac{1}{2}(z_i- \mu_{k_i})\Sigma_k^{-1}(z-\mu_k))}{\sqrt{(2 \pi)^{d}|\Sigma_k|}}\right) } + C \\
        =& d(z,\mu_k,\Sigma_k) + C \,,
\end{align*}
where $C$ is the posterior probability of all previous $N$ assignments.
Subbing this into \eqref{eq:difference-NLL-ent-objective1}, we get
\begin{gather*}
      d(z,\mu_k,\Sigma_k)  -  d(z,\mu_k,\Sigma_k) < \\
      \frac{1}{N+1} \left( \log(x_{k'} + 1) - \log (x_{k} + 1)\right)
      \iff \\
      ( d(z,\mu_{k},\Sigma_{k}) -  d(z,\mu_{k'},\Sigma_{k'})) < \frac{\lambda}{N+1} \left( \log(x_{k'} + 1) - \log (x_{k} + 1)\right) \iff \\
       d(z,\mu_{k},\Sigma_{k}) + \frac{\lambda}{N+1}\log{(x_{k} + 1)} <  d(z,\mu_{k},\Sigma_{k}) + \frac{\lambda}{N+1}\log{(x_{k} + 1)} \iff \\
       d(z,\mu_{k},\Sigma_{k}) + \frac{2\lambda}{N+1}\log{(x_{k} + 1)} <  d(z,\mu_{k'},\Sigma_{k'}) + \frac{2\lambda}{N+1}\log{(x_{k'} + 1)}\,. \label{eq:difference-NLL-ent-objective2}
\end{gather*}
Choosing pairwise between all $k,k'$ as per \eqref{eq:difference-NLL-ent-objective2} is equivalent to choosing $k$ so as to minimize 
\[
 d(z,\mu_{k},\Sigma_{k}) + \frac{2\lambda}{N+1}\log{(x_{k} + 1)} \,.
\]
\end{proof}

\begin{remark}
This shows that our proposed method closely approximates a maximization of the entropy of cluster labels. There is some similarity to those methods, discussed in Section \ref{sec:related-work}, that use an additional loss term to encourage greater entropy of soft assignments in each batch, but the important difference here is that we are maximizing the entropy of hard assignments. 
\end{remark}

\begin{theorem}
Assume that $N$ data points in a batch have already been assigned. Let $\mathcal{L}(k)$ be the batch likelihood, under the a $K$-component Gaussian mixture model, as described in Section \ref{subsec:comb-assignment}, after the $(N+1)$th data point is assigned to cluster $k$. Let $C$ and $B$ be, respectively, random variables indicating the cluster assignments in the batch and the batch indices. Then, the method presented in Section \ref{subsec:greedy-strategy} is equivalent, up to a small error term, to maximizing
\[
\log{\mathcal{L}(k)} + \lambda I(C;B)\,,
\]
for some $\lambda \in \mathbb{R}$ that does not depend on the cluster assignments in the batch.
\end{theorem}
\begin{proof}
By Lemma \ref{lemma:ent-diff}, the method in \eqref{eq:sum-objective} is equivalent to maximizing 
\begin{equation}
    \log{\mathcal{L}(k)} + \lambda H^k \,, \label{eq:equiv-to-NLL-ent-maximization}
\end{equation}
for $\lambda=\frac{1}{N+1}$. The mutual information $I(C;B)$ can be expressed in terms of entropy as 
\[
I(C;B) = H(C) - H(C|B)\,.
\]
Moreoever, we are making hard assignments so, given the cluster index, the distribution over cluster labels has all the probability on one cluster and has zero entropy. This means 
\[
I(C;B) = H(C) - H(C|B) = H(C) - 0 = H(C) \,.
\]
Subbing this into \eqref{eq:equiv-to-NLL-ent-maximization}, the result follows.
\end{proof}

\section{Full Imbalanced Results}

\FloatBarrier

\begin{table}[H]
\caption{Full imbalanced results, mean from 5 runs}
\resizebox{\textwidth}{!}{
\begin{tabular}{lllllllllllllllll}
\toprule
 &  & \multicolumn{3}{r}{ours} & \multicolumn{3}{r}{sk} & \multicolumn{3}{r}{ent} & \multicolumn{3}{r}{ss} & \multicolumn{3}{r}{ckm} \\
 &  & imb1 & imb2 & imb3 & imb1 & imb2 & imb3 & imb1 & imb2 & imb3 & imb1 & imb2 & imb3 & imb1 & imb2 & imb3 \\
\midrule
\multirow[t]{4}{*}{C10} & acc & 0.21 & 0.21 & 0.23 & 0.13 & 0.14 & 0.16 & 0.11 & 0.13 & 0.19 & 0.18 & 0.19 & 0.23 & 0.11 & 0.13 & 0.18 \\
 & nmi & 0.05 & 0.05 & 0.06 & 0.00 & 0.01 & 0.00 & 0.00 & 0.00 & 0.00 & 0.05 & 0.04 & 0.04 & 0.00 & 0.00 & 0.00 \\
 & ari & 0.00 & 0.49 & 0.38 & 0.30 & 0.02 & 0.01 & 0.03 & 0.00 & 0.00 & 0.00 & 0.09 & 0.12 & 0.14 & 0.00 & 0.00 \\
 & KL & 0.06 & 0.07 & 0.24 & 0.36 & 0.38 & 0.76 & 2.19 & 2.27 & 1.85 & 1.44 & 1.30 & 2.12 & 2.33 & 2.32 & 2.91 \\
\cline{1-17}
\multirow[t]{4}{*}{C100} & acc & 0.06 & 0.06 & 0.07 & 0.03 & 0.03 & 0.04 & 0.01 & 0.01 & 0.02 & 0.03 & 0.04 & 0.04 & 0.01 & 0.01 & 0.02 \\
 & nmi & 0.02 & 0.01 & 0.01 & 0.00 & 0.00 & 0.00 & 0.00 & 0.00 & 0.00 & 0.01 & 0.00 & 0.00 & 0.00 & 0.00 & 0.00 \\
 & ari & 0.05 & 0.05 & 0.06 & 0.00 & 0.01 & 0.00 & 0.00 & 0.00 & 0.00 & 0.05 & 0.04 & 0.04 & 0.00 & 0.00 & 0.00 \\
 & KL & 0.77 & 1.26 & 1.74 & 1.66 & 1.69 & 1.70 & 4.61 & 4.65 & 4.93 & 2.70 & 2.41 & 2.57 & 4.57 & 4.65 & 5.40 \\
\cline{1-17}
\multirow[t]{4}{*}{FM} & acc & 0.62 & 0.51 & 0.49 & 0.16 & 0.17 & 0.20 & 0.11 & 0.14 & 0.18 & 0.24 & 0.27 & 0.32 & 0.12 & 0.13 & 0.18 \\
 & nmi & 0.44 & 0.36 & 0.35 & 0.02 & 0.02 & 0.02 & 0.00 & 0.01 & 0.00 & 0.12 & 0.14 & 0.13 & 0.01 & 0.00 & 0.00 \\
 & ari & 0.02 & 0.01 & 0.01 & 0.00 & 0.00 & 0.00 & 0.00 & 0.00 & 0.00 & 0.01 & 0.00 & 0.00 & 0.00 & 0.00 & 0.00 \\
 & KL & 0.01 & 0.11 & 0.19 & 0.35 & 0.34 & 0.56 & 2.19 & 2.11 & 2.41 & 1.25 & 1.11 & 0.74 & 2.24 & 2.25 & 2.70 \\
\cline{1-17}
\multirow[t]{4}{*}{STL} & acc & 0.24 & 0.22 & 0.24 & 0.14 & 0.16 & 0.19 & 0.12 & 0.15 & 0.20 & 0.13 & 0.16 & 0.19 & 0.11 & 0.13 & 0.19 \\
 & nmi & 0.07 & 0.05 & 0.05 & 0.01 & 0.01 & 0.02 & 0.00 & 0.00 & 0.01 & 0.01 & 0.01 & 0.01 & 0.00 & 0.00 & 0.00 \\
 & ari & 0.44 & 0.36 & 0.35 & 0.02 & 0.02 & 0.02 & 0.00 & 0.01 & 0.00 & 0.12 & 0.14 & 0.13 & 0.01 & 0.00 & 0.00 \\
 & KL & 0.04 & 0.16 & 0.33 & 0.47 & 0.29 & 0.64 & 1.74 & 1.72 & 1.54 & 1.91 & 1.90 & 2.14 & 2.29 & 2.42 & 2.01 \\
\cline{1-17}
\multirow[t]{4}{*}{RD} & acc & 0.57 & 0.47 & 0.39 & 0.11 & 0.11 & 0.13 & 0.08 & 0.11 & 0.14 & 0.20 & 0.22 & 0.26 & 0.08 & 0.10 & 0.17 \\
 & nmi & 0.49 & 0.38 & 0.30 & 0.02 & 0.01 & 0.03 & 0.00 & 0.00 & 0.00 & 0.09 & 0.12 & 0.14 & 0.00 & 0.00 & 0.02 \\
 & ari & 0.07 & 0.05 & 0.05 & 0.01 & 0.01 & 0.02 & 0.00 & 0.00 & 0.01 & 0.01 & 0.01 & 0.01 & 0.00 & 0.00 & 0.00 \\
 & KL & 0.01 & 0.04 & 0.27 & 0.62 & 0.86 & 0.67 & 3.47 & 3.25 & 3.85 & 2.09 & 1.84 & 2.62 & 3.48 & 3.65 & 3.47 \\
\cline{1-17}
\bottomrule
\end{tabular}
}
\end{table}

\FloatBarrier

\section{Modified Variance Maximization}
As discussed in Section \ref{sec:experimental-eval}, one of the methods we compare to is that proposed by \cite{zhong2020deep}, which minimizes the sum of squares of the marginal soft assignments across a batch. The expectation of the square (and hence the sum of squares) can be decomposed as the square of the expectation plus the variance. Minimizing the sum of squares can then help to combat partition collapse as it involves minimizing the variance. However, empirically we find this method not to perform well, see Table \ref{tab:main-results}. Here, we show that a simply modified version of this method performs better than the original, though still less well than our method. Results are presented in Table \ref{tab:var-improved-comparison}.

The modification is to just minimize variance directly, rather than via sum of squares. Note that this may be equivalent in some formulations, if the probability of membership across clusters for a single data point is normalized to sum to $1$. Then the expectation of the sum of memberships is $1$, because it is $1$ deterministically, so the square of the expectation is also $1$ deterministically and, in particular, is independent of the cluster assignments. This means that minimizing the sum of squares with respect to cluster assignments, is identical to minimizing variance with respect to cluster assignments. In our model this is true. The probability of membership depends only on the distance to the cluster centroids, and is conditionally independent across clusters, given the cluster centroids. Details are not given in \cite{zhong2020deep} as to whether this holds in their method.

\begin{table}
\centering
\begin{tabular}{*{5}{c}}
\hline
 & & CA & Var & VarM \\\hline
	\multirow{4}{*}{Cifar10}  & Acc & \textbf{22.7 (2.07)} & 11.8 (1.72) & 20.8 (0.67) \\
	 & NMI & 10.1 (1.68) & 1.1 (1.15) & \textbf{11.2 (0.65)} \\
	 & ARI & 5.8 (0.93) & 0.3 (0.38) & \textbf{7.3 (0.50)} \\
	 & KL$^*$ & \textbf{0.04 (0.01)} & 1.32 (0.36) & 2.00 (0.09) \\
\hline
	\multirow{4}{*}{Cifar100}  & Acc & \textbf{6.4 (0.22)} & 1.2 (0.22) & 1.0 (0.00) \\
	 & NMI & \textbf{13.2 (0.37)} & 0.6 (1.05) & 0.0 (0.00) \\
	 & ARI & \textbf{1.7 (0.14)} & 0.0 (0.04) & 0.0 (0.00) \\
    & KL$^*$ & \textbf{0.81 (0.07)} & 3.76 (0.37) & 6.64 (0.00) \\
\hline
	\multirow{4}{*}{FashionMNIST}  & Acc & \textbf{54.5 (6.96)} & 10.0 (0.04) & 37.4 (2.53) \\
	 & NMI & \textbf{53.2 (4.23)} & 0.0 (0.04) & 42.8 (2.08) \\
	 & ARI & \textbf{39.1 (6.29)} & 0.0 (0.00) & 27.1 (1.78) \\
    & KL$^*$ &  \textbf{0.04 (0.00)} & 1.34 (0.57) & 1.70 (0.07) \\
\hline
	\multirow{4}{*}{STL}  & Acc & \textbf{23.5 (1.42)} & 10.1 (0.20) & 22.6 (1.52) \\
	 & NMI & \textbf{13.7 (1.33)} & 0.0 (0.08) & 11.4 (1.70) \\
	 & ARI & 7.1 (0.70) & 0.0 (0.00) & \textbf{7.1 (1.16)} \\
    & KL$^*$ &  \textbf{0.09 (0.01)} & 2.23 (0.12) & 1.50 (0.13) \\
\hline
\end{tabular}
\caption{Comparison between the modified variance minimization method, denoted `VarM', the original variance minimization method from \cite{zhong2020deep}, denoted `Var', and our method, denoted `CA'.}
\label{tab:var-improved-comparison}
\end{table}

\section{Calcuation of Entropies of Matrices}
Let $h,s: \mathbb{R}^{4 \times 3} \rightarrow \mathbb{R}^3$ be the functions that compute the marginal hard and soft cluster distributions for a given matrix of batch assignment probabilities. Then, for matrices
\[
D_1 = \begin{bmatrix}
.98 & .01 & .01 \\
.98 & .01 & .01 \\
.49 & .50 & .01 \\
.49 & .01 & .50 
\end{bmatrix}
D_2 = \begin{bmatrix}
.34 & .33 & .33 \\
.34 & .33 & .33 \\
.34 & .33 & .33 \\
.34 & .33 & .33 
\end{bmatrix}  \,,
\]
we have
\begin{align*}
    &s(D_1) = \begin{bmatrix}
.74, .13, .13 
\end{bmatrix} &H(h(D_1)) = 1.10 \\
    &h(D_1) = \begin{bmatrix}
.5,.25,.25 
\end{bmatrix} &H(s(D_1)) = 1.50 \\
     &s(D_2) =\begin{bmatrix}
.34.,.33,.33 
\end{bmatrix} &H(s(D_2)) = 1.58 \\
    &h(D_2) = \begin{bmatrix}
1,0,0 
\end{bmatrix} &H(h(D_2)) = 0 \,.
\end{align*}

\section{Explicit Distributions in Imbalanced Experiments}
For the three imbalanced settings, imb. 1, imb. and imb. 3, the relative frequences of clusters are as follows:
imb 1.: $1.0,1.0,1.0,1.0,1.0,1.0,0.95,0.9,0.85,0.8$. \\
imb 2.: $1.0,0.95,0.9,0.85,0.8,0.75,0.7,0.65,0.6,0.55$. \\
imb 3.: $1.0,0.9,0.8,0.7,0.6,0.5,0.4,0.3,0.2,0.1$.

\end{document}


\section{train.py}
from time import time
import torch.nn.functional as F
from sklearn.model_selection import train_test_split
from sklearn.linear_model import LogisticRegression
from sklearn.neighbors import KNeighborsClassifier
from os.path import join
from dl_utils.label_funcs import label_counts, get_trans_dict, accuracy
from dl_utils.tensor_funcs import cudify, numpyify
from dl_utils.misc import set_experiment_dir, asMinutes, scatter_clusters
from dl_utils.torch_misc import CifarLikeDataset
import torch.nn as nn
import torch.optim as optim
from torch.utils.data import DataLoader
from torch.distributions import Categorical
import cl_args
from pdb import set_trace
import numpy as np
import torch
from sklearn.metrics import normalized_mutual_info_score, adjusted_rand_score

class ClusterNet(nn.Module):
    def __init__(self,ARGS):
        super().__init__()
        self.bs_train = ARGS.batch_size_train
        self.bs_val = ARGS.batch_size_val
        self.nc = ARGS.nc
        self.nz = ARGS.nz
        #self.sigma = ARGS.sigma
        if ARGS.dataset == 'tweets':
            counts = np.load('datasets/tweets/cluster_label_counts.npy')
            self.log_prior = np.log(counts/counts.sum())
        else:
            self.prior = ARGS.prior
        self.log_prior = np.log(self.prior)

        if ARGS.dataset == 'imt':
            out_conv_shape = 13
        elif ARGS.dataset == 'stl':
            out_conv_shape = 21
        elif ARGS.dataset == 'fashmnist':
            out_conv_shape = 4
        else:
            out_conv_shape = 5
        nc = 1 if ARGS.dataset == 'fashmnist' else 3
        self.conv1 = nn.Conv2d(3, 6, 5)
        if ARGS.arch == 'alex':
            self.net = torch.hub.load('pytorch/vision:v0.10.0', 'alexnet', pretrained=False)
            self.net.classifier = self.net.classifier[:5] # remove final linear and relu
            self.net.classifier[4] = nn.Linear(4096,self.nz,device='cuda')
        if ARGS.arch == 'res':
            self.net = torch.hub.load('pytorch/vision:v0.10.0', 'resnet18', pretrained=False)
            self.net.fc.weight.data = self.net.fc.weight.data[:self.nz]
            self.net.fc.bias.data = self.net.fc.bias.data[:self.nz]
            self.net.fc.out_features = self.nz
        elif ARGS.arch == 'simp':
            self.net = nn.Sequential(
                nn.Conv2d(nc, 6, 5),
                nn.BatchNorm2d(6),
                nn.ReLU(),
                nn.MaxPool2d(2, 2),
                nn.Conv2d(6, 16, 5),
                nn.BatchNorm2d(16),
                nn.ReLU(),
                nn.MaxPool2d(2, 2),
                nn.Flatten(1),
                nn.Linear(16 * out_conv_shape * out_conv_shape, self.nz),
                )
        elif ARGS.arch == 'fc':
            self.net = nn.Sequential(
                nn.Linear(768,ARGS.hidden_dim),
                nn.ReLU(),
                nn.Linear(ARGS.hidden_dim,self.nz))

        self.opt = optim.Adam(self.net.parameters(),lr=ARGS.lr)

        self.centroids = torch.randn(ARGS.nc,ARGS.nz,requires_grad=True,device='cuda',dtype=torch.double)
        self.inv_covars = (1/ARGS.sigma)*torch.eye(ARGS.nz,requires_grad=ARGS.is_train_covars,device='cuda',dtype=torch.double).repeat(self.nc,1,1)
        self.half_log_det_inv_covars = torch.log(torch.tensor(ARGS.sigma))*self.nz/2
        #self.inv_covars = torch.randn(ARGS.nc,ARGS.nz,ARGS.nz,requires_grad=True,device='cuda',dtype=torch.double)
        self.ng_opt = optim.Adam([{'params':self.centroids}],lr=ARGS.lr)
        if ARGS.is_train_covars:
            self.ng_opt.add_param_group({'params':self.inv_covars})
        self.cluster_log_probs = None
        self.cluster_counts = torch.zeros(ARGS.nc,device='cuda').long()
        self.raw_counts = torch.zeros(ARGS.nc,device='cuda').long()
        self.total_soft_counts = torch.zeros(ARGS.nc,device='cuda')

        self.epoch_num = -1
        self.temp = ARGS.temp

        self.training = True

    def train(self):
        self.training = True
        self.net.train()
        self.centroids.requires_grad = True

    def eval(self):
        self.training = False
        self.net.eval()
        self.centroids.requires_grad = False

    def reset_scores(self):
        self.cluster_counts = torch.zeros(self.nc,device='cuda').long()
        self.raw_counts = torch.zeros(self.nc,device='cuda').long()

    def init_keys_as_dpoints(self,dloader):
        self.eval()
        inp,targets = next(iter(dloader))
        inp = inp[:self.nc]
        while len(inp) < self.nc:
            new_inp,targets = next(iter(dloader))
            inp = torch.cat([inp,new_inp[:self.nc-len(inp)]])
        sample_feature_vecs = self.net(inp.cuda())
        self.centroids = sample_feature_vecs.clone().detach().double().requires_grad_(True)

    def forward(self, inp):
        self.bs = inp.shape[0]
        feature_vecs = self.net(inp)
        cluster_dists = feature_vecs[:,None]-self.centroids
        self.cluster_log_probs = torch.einsum('bcu,czu,bcz->bc',cluster_dists,self.inv_covars,cluster_dists) - self.half_log_det_inv_covars
        #assert torch.allclose(self.cluster_log_probs, cluster_dists.norm(dim=2)**2)
        #assert all([torch.allclose(self.cluster_log_probs[b,c],(cluster_dists[b,c]@self.inv_covars[c]) @cluster_dists[b,c]) for b in range(self.bs) for c in range(self.nc)])
        self.assign_batch()
        return feature_vecs

    def assign_batch(self):
        if ARGS.ng:
            self.cluster_loss,self.batch_assignments = neural_gas_loss(.1*self.cluster_log_probs+(self.cluster_counts+1).log(),self.temp)
        elif ARGS.var:
            self.assign_batch_var()
        elif ARGS.kl:
            self.assign_batch_kl()
        elif ARGS.sinkhorn:
            self.assign_batch_sinkhorn()
        elif ARGS.no_reg:
            min_dists, self.batch_assignments = self.cluster_log_probs.min(axis=1)
            self.cluster_loss = min_dists.mean()
        else:
            self.assign_batch_probabilistic()

        self.raw_counts.index_put_(indices=[self.cluster_log_probs.argmin(axis=1)],values=torch.ones_like(self.batch_assignments),accumulate=True)
        if ARGS.ng or ARGS.sinkhorn or ARGS.kl:
            self.cluster_counts.index_put_(indices=[self.batch_assignments],values=torch.ones_like(self.batch_assignments),accumulate=True)
        if not ARGS.sinkhorn or ARGS.ng:
            self.soft_counts = (-self.cluster_log_probs).softmax(axis=1).sum(axis=0).detach()
        self.total_soft_counts += self.soft_counts

    def assign_batch_sinkhorn(self):
        """Implements the method referred to in the paper as 'SK'"""
        with torch.no_grad():
            #hard_counts = (torch.arange(self.nc).cuda() == self.cluster_log_probs.argmax(axis=1,keepdims=True)).float()
            soft_assignments = sinkhorn(-self.cluster_log_probs,is_hard_reg=ARGS.hard_sinkhorn,eps=.5,niters=15)
        if self.prior != 'uniform' and self.epoch_num>0 and self.batch_assignments = soft_assignments.argmin(axis=1)
        self.soft_counts = soft_assignments.sum(axis=0).detach()
        self.cluster_loss = self.cluster_log_probs[torch.arange(self.bs),self.batch_assignments].mean()

    def assign_batch_var(self):
        """Implements the method referred to in the paper as 'VAR'"""
        self.batch_assignments = self.cluster_log_probs.argmin(axis=1)
        #self.cluster_loss = 10*(self.cluster_log_probs**2).sum()
        if ARGS.var_improved:
            self.cluster_loss = 10*self.cluster_log_probs.mean(axis=0).var()
        else:
            self.cluster_loss = 10*(self.cluster_log_probs.mean(axis=0)**2).mean()
        self.cluster_loss +=.1*self.cluster_log_probs[torch.arange(self.bs),self.batch_assignments].mean()

    def assign_batch_kl(self):
        """Implements the method referred to in the paper as 'ENT'"""
        self.batch_assignments = self.cluster_log_probs.argmin(axis=1)
        self.cluster_loss = 100*-Categorical(self.cluster_log_probs.mean(axis=0)).entropy()
        if ARGS.kl_cent:
            self.cluster_loss +=.1*self.cluster_log_probs[torch.arange(self.bs),self.batch_assignments].mean()
        else:
            self.cluster_loss += .1*Categorical(self.cluster_log_probs).entropy().mean()

    def assign_batch_probabilistic(self):
        """Implements the our method, referred to in the paper as 'CA'"""
        assigned_key_order = []
        cost_table = self.cluster_log_probs.transpose(0,1).flatten(1).transpose(0,1)
        self.batch_assignments = torch.zeros_like(self.cluster_log_probs[:,0]).long()
        unassigned_idxs = torch.ones_like(cost_table[:,0]).bool()
        #cost_table = flat_x/(2*ARGS.sigma)
        if ARGS.imbalance > 0 and self.epoch_num > 0:
            cost_table -= self.translated_log_prior
        had_repeats = False
        if not self.training and not ARGS.constrained_eval:
            self.batch_assignments = self.cluster_log_probs.argmin(axis=1)
            return
        assign_iter = 0
        while unassigned_idxs.any():
            assert (~unassigned_idxs).sum() == assign_iter or had_repeats
            cost = (cost_table[unassigned_idxs]+(self.cluster_counts+1).log()).min()
            nzs = ((cost_table+(self.cluster_counts+1).log() == cost)*unassigned_idxs[:,None]).nonzero()
        """Implements the our method, referred to in the paper as 'CA'"""
            if len(nzs)!=1: had_repeats = True
            new_vec_idx, new_assigned_key = nzs[0]
            assert unassigned_idxs[new_vec_idx]
            unassigned_idxs[new_vec_idx] = False
            assigned_key_order.append(new_vec_idx)
            self.batch_assignments[new_vec_idx] = new_assigned_key
            self.cluster_counts[new_assigned_key] += 1
            #assert cost > 0
            assign_iter += 1
        self.cluster_loss = cost_table[torch.arange(self.bs),self.batch_assignments].mean()

    def train_one_epoch(self,trainloader):
        self.train()
        running_loss = 0.0
        self.reset_scores()
        for i, data in enumerate(trainloader):
            if not ARGS.keep_scores:
                self.reset_scores()
            self.batch_inputs, self.batch_labels = data
            if i==ARGS.db_at: set_trace()
            self(self.batch_inputs.cuda())
            self.cluster_loss.backward()
            self.opt.step()
            self.ng_opt.step()
            self.opt.zero_grad(); self.ng_opt.zero_grad()
            if i \% 10 == 0 and i > 0:
                if ARGS.track_counts:
                    for k,v in enumerate(self.cluster_counts):
                        if (rc := self.raw_counts[k].item()) == 0:
                            continue
                        print(f"{k} constrained: {v.item()}\traw: {self.raw_counts[k].item()}\tsoft: {self.soft_counts[k].item():.3f}")
                if not ARGS.suppress_prints:
                    print(f'batch index: {i}\tloss: {running_loss/10:.3f}')
                running_loss = 0.0
            running_loss += self.cluster_loss.item()
            if (self.centroids==0).all(): set_trace()
            if ARGS.is_test > 0:
                break

    def train_epochs(self,num_epochs,dset,val_too=True):
        trainloader = DataLoader(dset,batch_size=self.bs_train,shuffle=True,num_workers=8)
        testloader = DataLoader(dset,batch_size=self.bs_val,shuffle=False,num_workers=8)
        best_acc = -1
        best_nmi = -1
        best_ari = -1
        best_kl_star = -1
        best_linear_probe_acc = -1
        best_knn_probe_acc = -1
        if ARGS.warm_start:
            self.init_keys_as_dpoints(trainloader)
        for epoch_num in range(num_epochs):
            self.epoch_num = epoch_num
            self.total_soft_counts = torch.zeros_like(self.total_soft_counts)
            self.train_one_epoch(trainloader)
            if val_too:
                self.total_soft_counts = torch.zeros_like(self.total_soft_counts)
                with torch.no_grad():
                    self.test_epoch_unsupervised(testloader)
                model_distribution = self.epoch_hard_counts/self.epoch_hard_counts.sum()
                log_quot = np.log((model_distribution/self.prior)+1e-8)
                self.kl_star = np.dot(model_distribution,log_quot)
                if self.nmi > best_nmi:
                    best_nmi = self.nmi
                    best_acc = self.acc
                    best_ari = self.ari
                    best_kl_star = self.kl_star
                else:
                    with torch.no_grad():
                        self.test_epoch_unsupervised(testloader)
                linear_probe_acc, knn_probe_acc = self.train_test_probes(dset)
                if linear_probe_acc > best_linear_probe_acc:
                    best_linear_probe_acc = linear_probe_acc
                    best_knn_probe_acc = knn_probe_acc
        print(f"Best Acc: {best_acc:.3f}\tBest NMI: {best_nmi:.3f}\tBest ARI: {best_ari:.3f}\tBest KL*:{best_kl_star:.5f}\tBest linear probe acc:{best_linear_probe_acc:.3f}\tBest KNN probe acc:{best_knn_probe_acc:.3f}")
        with open(join(ARGS.exp_dir,'ARGS.txt'),'w') as f:
            f.write(f'Dataset: {ARGS.dataset}\n')
            for a in ['batch_size_train','nz','hidden_dim','lr','sigma']:
                f.write(f'{a}: {getattr(ARGS,a)}\n')
            f.write(f'warm_start: {ARGS.warm_start}\n')

        with open(join(ARGS.exp_dir,'results.txt'),'w') as f:
            f.write(f'ACC: {best_acc:.3f}\nNMI: {best_nmi:.3f}\nARI: {best_ari:.3f}\n')
            f.write(f'KL-star: {best_kl_star:.3f}\nLin-Acc: {best_linear_probe_acc:.3f}\nKNN-Acc: {best_knn_probe_acc:.3f}\n')

    def train_test_probes(self,dset):
        self.eval()
        dloader = DataLoader(dset,batch_size=self.bs_val,shuffle=False,num_workers=8)
        all_encodings = []
        for i,data in enumerate(dloader):
            images, labels = data
            encodings = numpyify(self.net(images.cuda()))
            all_encodings.append(encodings)
        X = np.concatenate(all_encodings)
        y = dset.targets
        X_tr,X_ts,y_tr,y_ts = train_test_split(X,y,test_size=0.33)
        lin_reg = LogisticRegression().fit(X_tr,y_tr)
        lin_test_preds = lin_reg.predict(X_ts)
        lin_test_acc = (lin_test_preds==y_ts).mean()
        knn_reg = KNeighborsClassifier(n_neighbors=ARGS.n_neighbors).fit(X_tr,y_tr)
        knn_test_preds = knn_reg.predict(X_ts)
        knn_test_acc = (knn_test_preds==y_ts).mean()
        return lin_test_acc, knn_test_acc

    def test_epoch_unsupervised(self,testloader):
        self.eval()
        preds = []
        all_feature_vecs = []
        data_for_clusters = [[] for _ in range(self.nc)]
        for images,labels in testloader:
            feature_vecs = self(images.cuda())
            all_feature_vecs.append(numpyify(feature_vecs))
            assignments =self.batch_assignments
            for cluster_idx in range(self.nc):
                data_for_clusters[cluster_idx].append(feature_vecs[assignments==cluster_idx])
            preds.append(assignments.detach().cpu().numpy())
        pred_array = np.concatenate(preds)
        num_of_each_label = label_counts(pred_array)
        self.epoch_hard_counts = np.zeros(self.nc)
        for ass,num in num_of_each_label.items():
            self.epoch_hard_counts[ass] = num
        if ARGS.estimate_covars and len(num_of_each_label) == self.nc: # don't set covars if some dpoints missing
            unnormed_inv_covars = torch.stack([torch.inverse(torch.cat(cd).T.cov()) for cd in data_for_clusters])
            self.inv_covars = (unnormed_inv_covars*self.inv_covars.mean()/unnormed_inv_covars.mean()).double()
            if unnormed_inv_covars.isnan().any():
                breakpoint()
        self.epoch_soft_counts = self.total_soft_counts.detach().cpu().numpy()
        self.gt = testloader.dataset.targets
        self.trans_dict = get_trans_dict(np.array(self.gt),pred_array)
        self.acc = accuracy(pred_array,np.array(self.gt))
        if self.acc == 0:
            breakpoint()
        idx_array = np.array(list(self.trans_dict.keys())[:-1])
        self.translated_prior = self.prior[idx_array]
        self.translated_log_prior = cudify(self.log_prior[idx_array])
        self.nmi = normalized_mutual_info_score(pred_array,np.array(self.gt))
        self.ari = adjusted_rand_score(pred_array,np.array(self.gt))
        self.hcv = self.epoch_hard_counts.var()/self.epoch_hard_counts.mean()
        self.scv = self.epoch_soft_counts.var()/self.epoch_hard_counts.mean()
        if ARGS.viz_clusters:
            feature_vecs_array = np.concatenate(all_feature_vecs)
            import umap
            to_viz = umap.UMAP().fit_transform(feature_vecs_array)
            ax = scatter_clusters(to_viz,testloader.dataset.targets)
            breakpoint()

def neural_gas_loss(v,temp):
    n_instances, n_clusters = v.shape
    weightings = (-torch.arange(n_clusters,device=v.device)/temp).exp()
    sorted_v, assignments_order = torch.sort(v)
    assert (sorted_v**2 * weightings).mean() < ((sorted_v**2).mean() * weightings.mean())
    return (sorted_v**2 * weightings).sum(axis=1), assignments_order[:,0]

def sinkhorn(scores, is_hard_reg=False, eps=0.05, niters=3):
    Q = torch.exp(scores / eps).T
    #Q = torch.softmax(scores / eps,1).T
    Q /= sum(Q)
    eps2 = 0.1
    if is_hard_reg:
        hard_counts = (torch.arange(Q.shape[0]).cuda()[:,None] == Q.argmax(axis=0)).float()
        #Q = torch.cat([Q,1*hard_counts.sum(axis=1,keepdims=True)],axis=1)
        Q = (Q + eps2*hard_counts) / (1+eps2)
    K, B = Q.shape
    r, c = torch.ones(K,device=Q.device) / K, torch.ones(B,device=Q.device) / B
    for _ in range(niters):
        u = torch.sum(Q, dim=1)
        Q *= (r / u).unsqueeze(1)
        Q *= (c / torch.sum(Q, dim=0)).unsqueeze(0)
    if is_hard_reg:
        Q = (Q*2) - hard_counts
        #Q = Q[:,:-1]
    return (Q / torch.sum(Q, dim=0, keepdim=True)).T

if __name__ == '__main__':
    ARGS,dataset = cl_args.get_cl_args_and_dset()
    ARGS.exp_dir = set_experiment_dir(f'experiments/{ARGS.expname}',name_of_trials='experiments/tmp',overwrite=ARGS.overwrite)
    start_time = time()
    cluster_net = ClusterNet(ARGS).cuda()
    cluster_net.train_epochs(ARGS.epochs,dataset,val_too=True)
    print(f'Total time: {asMinutes(time()-start_time)}')

\section{cl_args.py}
import math
import argparse
import torch
from dl_utils.torch_misc import CifarLikeDataset
import numpy as np
import get_datasets
from HAR.make_dsets import StepDataset

RELEVANT_ARGS = []
def get_cl_args():
    parser = argparse.ArgumentParser()
    train_type_group = parser.add_mutually_exclusive_group()
    train_type_group.add_argument('--kl',action='store_true')
    train_type_group.add_argument('--var',action='store_true')
    train_type_group.add_argument('--ng',action='store_true')
    train_type_group.add_argument('--no_reg',action='store_true')
    train_type_group.add_argument('--no_cluster_loss',action='store_true')
    train_type_group.add_argument('--sinkhorn',action='store_true')
    parser.add_argument('--arch',type=str,choices=['alex','res','simp','fc','1dcnn'],default='simp')
    parser.add_argument('--gpu',type=str,default='0')
    parser.add_argument('--batch_size_train',type=int,default=256)
    parser.add_argument('--batch_size_val',type=int,default=1024)
    parser.add_argument('--constrained_eval',action='store_true')
    parser.add_argument('--db_at',type=int,default=-1)
    parser.add_argument('--estimate_covars',action='store_true')
    parser.add_argument('--expname',type=str,default='tmp')
    parser.add_argument('--ckm',action='store_true')
    parser.add_argument('--hard_sinkhorn',action='store_true')
    parser.add_argument('--help_sinkhorn',action='store_true')
    parser.add_argument('--hidden_dim',type=int,default=512)
    parser.add_argument('--imbalance',type=int,default=0)
    parser.add_argument('--is_train_covars',action='store_true')
    parser.add_argument('--keep_scores',action='store_true')
    parser.add_argument('--kl_cent',action='store_true')
    parser.add_argument('--linear_probe',action='store_true')
    parser.add_argument('--lr',type=float,default=1e-3)
    parser.add_argument('--n_neighbors',type=int,default=10)
    parser.add_argument('--nc',type=int,default=10)
    parser.add_argument('--nz',type=int,default=128)
    parser.add_argument('--overwrite',action='store_true')
    parser.add_argument('--pretrain_frac',type=float,default=0.5)
    parser.add_argument('--sigma',type=float,default=100.)
    parser.add_argument('--soft_train',action='store_true')
    parser.add_argument('--suppress_prints',action='store_true')
    parser.add_argument('--temp',type=float,default=1.)
    parser.add_argument('--is_test','-t',action='store_true')
    parser.add_argument('--track_counts',action='store_true')
    parser.add_argument('--var_improved',action='store_true')
    parser.add_argument('--verbose',action='store_true')
    parser.add_argument('--viz_clusters',action='store_true')
    parser.add_argument('--warm_start',action='store_true')
    parser.add_argument('-d','--dataset',type=str,choices=['imt','c10','c100','svhn','stl','fashmnist','tweets','realdisp'],default='c10')
    parser.add_argument('-e','--epochs',type=int,default=1)
    ARGS = parser.parse_args()
    if ARGS.is_test > 0:
        ARGS.expname = 'tmp'
    return ARGS

def make_dset_imbalanced(dset,nc,class_probs):
    imbalanced_data = []
    imbalanced_targets = []
    for i,p in enumerate(class_probs):
        targets = np.array(dset.targets)
        label_mask = targets==i
        rand_mask =np.random.rand(sum(label_mask))<p # select each independently, roughly get 1/p
        new_data = dset.data[label_mask][rand_mask]
        new_targets = targets[label_mask][rand_mask]
        imbalanced_data.append(new_data)
        imbalanced_targets.append(new_targets)
    imbalanced_data_arr = np.concatenate(imbalanced_data)
    imbalanced_targets_arr = np.concatenate(imbalanced_targets)
    assert len(imbalanced_data_arr) == len(imbalanced_targets_arr)
    return CifarLikeDataset(imbalanced_data_arr,imbalanced_targets_arr,transform=dset.transform)

def make_dset_imbalanced_har(dset,nc,class_probs):
    chunked_data = np.stack([dset.data[dset.step_size*i:(dset.step_size*i)+dset.window_size] for i in range(len(dset.targets))])
    chunked_data = np.expand_dims(chunked_data,1)
    chunked_dset = CifarLikeDataset(chunked_data,dset.targets)
    return make_dset_imbalanced(chunked_dset,nc,class_probs)

def get_cl_args_and_dset():
    args = get_cl_args()

    dataset = get_datasets.get_dset(args.dataset,args.is_test)
    n_classes = len(set(dataset.targets))
    if args.imbalance==1:
        n = n_classes//2
        m = n_classes - n
        class_probs=np.concatenate([np.ones(m),1-0.2*np.linspace(0,1,n)])
    elif args.imbalance==2:
        class_probs = 1-0.5*np.linspace(0,1-1/n_classes,n_classes)
    elif args.imbalance==3:
        class_probs = 1-np.linspace(0,1-1/n_classes,n_classes)
    if args.dataset == 'c100':
        args.nc = 100
    elif args.dataset == 'imt':
        if args.imbalance>0:
            class_probs = np.tile(class_probs,20)
        args.nc = 200
    elif args.dataset == 'tweets':
        args.nc = 269
        args.arch = 'fc'
    elif args.dataset == 'realdisp':
        args.nc = 33
        args.nz = 32
        args.arch = '1dcnn'
    else:
        args.nc = 10

    is_har = args.dataset == 'realdisp'
    if args.imbalance > 0:
        imb_dset_func = make_dset_imbalanced_har if is_har else make_dset_imbalanced
        dataset = imb_dset_func(dataset,args.nc,class_probs)
        args.prior = class_probs/class_probs.sum()
    else:
        args.prior = np.ones(args.nc)/args.nc
    return args, dataset

\section{get_datasets.py}
import torch
from os.path import join
import torchvision
from torch.utils import data
from torchvision.transforms import Compose, Normalize, ToTensor
import numpy as np
from functools import partial
from dl_utils.torch_misc import CifarLikeDataset
from dl_utils.tensor_funcs import numpyify
from HAR.make_dsets import make_realdisp_dset
from HAR.project_config import realdisp_info

def get_tweets(is_use_testset):
    X = np.load('datasets/tweets/roberta_doc_vecs.npy')
    y = np.load('datasets/tweets/cluster_labels.npy')
    return CifarLikeDataset(X,y)

def get_cifar10(is_use_testset):
    transform = Compose([ToTensor(),Normalize((0.5,0.5,0.5), (0.5,0.5,0.5))])
    data_dir = '~/dataset/cifar10_data'
    dset = torchvision.datasets.CIFAR10(root=data_dir,train=not is_use_testset,transform=transform,download=True)
    return dset.data, dset.targets

def get_cifar100(is_use_testset):
    transform = Compose([ToTensor(),Normalize((0.4914, 0.4822, 0.4465), (0.2675, 0.2565, 0.2761))])
    data_dir = '~/datasets/cifar100_data'
    dset = torchvision.datasets.CIFAR100(root=data_dir,train=not is_use_testset,transform=transform,download=True)
    return dset.data, dset.targets

def get_fashmnist(is_use_testset):
    transform = Compose([ToTensor()])
    data_dir = '~/dataset/fashmnist_data'
    dset = torchvision.datasets.FashionMNIST(root=data_dir,train=not is_use_testset,transform=transform,download=True)
    return dset.data, dset.targets

def get_stl(is_test_run):
    transform = Compose([ToTensor(),Normalize((0.5, 0.5, 0.5), (0.5, 0.5, 0.5))])
    data_dir = './dataset/stl_data'
    def wrapper(dset_func):
        def inner():
            dset = dset_func()
            return CifarLikeDataset(np.transpose(dset.data,(0,3,2,1)),dset.labels,transform)
        return inner
    if is_test_run:
        dset = torchvision.datasets.STL10(root=data_dir,split='train',transform=transform,download=True)
    else:
        dset = torchvision.datasets.STL10(root=data_dir,split='test',transform=transform,download=True)

    return np.transpose(dset.data,(0,3,2,1)), dset.labels

def get_train_or_test_dset(dset_name,is_use_testset):
    if dset_name=='c10':
        X,y = get_cifar10(is_use_testset)
    elif dset_name=='c100':
        X,y = get_cifar100(is_use_testset)
    elif dset_name=='stl':
        X,y = get_stl(True)
    elif dset_name=='fashmnist':
        X,y = get_fashmnist(is_use_testset)
    elif dset_name=='imt':
        X,y = get_imagenet_tiny(is_use_testset)
    else:
        print(f'\nUNRECOGNIZED DATASET: {dset_name}\n')
    X = numpyify(X)
    y = numpyify(y)
    return X, y

def get_dset(dset_name,is_test_run):
    if dset_name=='realdisp':
        subj_ids = realdisp_info().possible_subj_ids
        if is_test_run:
            subj_ids = subj_ids[:1]
        dset,_ = make_realdisp_dset(step_size=5,window_size=512,subj_ids=subj_ids)
        return dset
    else:
        X, y = get_train_or_test_dset(dset_name,True)
    if is_test_run:
        X = X[:1000]
        y = y[:1000]
    elif dset_name!='imt': # no train-test split in im-tiny
        X_tr, y_tr = get_train_or_test_dset(dset_name,False)
        X = np.concatenate([X_tr,X])
        y = np.concatenate([y_tr,y])
    if dset_name == 'fashmnist':
        transform = ToTensor()
    elif dset_name == 'imt':
        transform = None
    else:
        transform = Compose([ToTensor(),Normalize((0.5, 0.5, 0.5), (0.5, 0.5, 0.5))])
    return CifarLikeDataset(X,y,transform=transform)

def get_imagenet_tiny(test):
    data_for_each_class = []
    labels_for_each_class = []
    for class_idx in range(200):
        np_class_data = np.load(join('tiny-imagenet-200/np_data',f'{class_idx}.npy'))
        np_class_labels = np.load(join('tiny-imagenet-200/np_data',f'{class_idx}_labels.npy'))
        if test == 1:
            np_class_data = np_class_data[:50]
            np_class_labels = np_class_labels[:50]
        elif test == 2:
            np_class_data = np_class_data[:20]
            np_class_labels = np_class_labels[:20]
        data_for_each_class.append(np_class_data)
        labels_for_each_class.append(np_class_labels)

    data_as_array = torch.tensor(np.concatenate(data_for_each_class)).transpose(1,3).float()
    labels_as_array = torch.tensor(np.concatenate(labels_for_each_class)).long()

    return data_as_array,labels_as_array

\section{utils.py}
import torch.nn as nn
import math
import torch

class EncByLayer(nn.Module):
    def __init__(self,ncvs,ksizes,strides,paddings,max_pools,show_shapes):
        super(EncByLayer,self).__init__()
        self.show_shapes = show_shapes
        num_layers = len(ksizes)
        assert all([isinstance(x,int) for l in (ncvs,ksizes,strides) for x in l])
        assert all(len(x)==num_layers for x in (ksizes,strides,max_pools))
        conv_layers = []
        for i in range(num_layers):
            if i<num_layers-1:
                conv_layer = nn.Sequential(
                nn.Conv2d(ncvs[i],ncvs[i+1],ksizes[i],strides[i],paddings[i]),
                nn.BatchNorm2d(ncvs[i+1]),
                nn.LeakyReLU(0.3),
                nn.MaxPool2d(max_pools[i])
                )
            else: #No batch norm on the last layer
                conv_layer = nn.Sequential(
                nn.Conv2d(ncvs[i],ncvs[i+1],ksizes[i],strides[i],paddings[i]),
                nn.LeakyReLU(0.3),
                nn.MaxPool2d(max_pools[i])
                )
            conv_layers.append(conv_layer)
        self.conv_layers = nn.ModuleList(conv_layers)

    def forward(self,x):
        if self.show_shapes: print(x.shape)
        for i,conv_layer in enumerate(self.conv_layers):
            x = conv_layer(x)
            # sometimes errors in the batchnorm if size is already down to 1
            if self.show_shapes: print(i,x.shape)
        return x

class DecByLayer(nn.Module):
    def __init__(self,ncvs,ksizes,strides,paddings,show_shapes):
        super(DecByLayer,self).__init__()
        self.show_shapes = show_shapes
        n_layers = len(ksizes)
        assert all([isinstance(x,int) for l in (ncvs,ksizes,strides,paddings) for x in l])
        assert all(len(x)==n_layers for x in (ksizes,strides,paddings) )
        conv_trans_layers = [nn.Sequential(
                nn.ConvTranspose2d(ncvs[i],ncvs[i+1],ksizes[i],strides[i],paddings[i]),
                nn.BatchNorm2d(ncvs[i+1]),
                nn.LeakyReLU(0.3),
                )
            for i in range(n_layers)]
        self.conv_trans_layers = nn.ModuleList(conv_trans_layers)

    def forward(self,x):
        if self.show_shapes: print(x.shape)
        for i,conv_trans_layer in enumerate(self.conv_trans_layers):
            x = conv_trans_layer(x)
            if self.show_shapes: print(i,x.shape)
        return x

def increment_approx_exponentially(insize,outsize,n_increments):
    base = (outsize/insize)**(1/(n_increments-1))
    assert base > 1
    increments = [int(insize*base**(i)) for i in range(n_increments)]
    if increments[-1] != outsize:
        print(f'readjusting output size from {increments[-1]} to {outsize}')
        increments[-1] = outsize
    if increments[0] != insize:
        breakpoint()
    return increments

def build_convt_net(in_chans,in_shape,outsize,n_layers):
    chans = list(reversed(increment_approx_exponentially(in_chans,outsize,n_layers+1)))
    sizes = increment_approx_exponentially(1,in_shape,n_layers+1)
    ksizes,strides,paddings = zip(*[infer_single_layer_shape(sizes[i+1],sizes[i]) for i in range(len(sizes)-1)])
    return DecByLayer(chans,ksizes,strides,paddings,False)

def build_conv_net(in_chans,in_shape,outsize,n_layers):
    chans = increment_approx_exponentially(in_chans,outsize,n_layers+1)
    sizes = increment_approx_exponentially(1,in_shape,n_layers+1) # no flatten
    ksizes,strides,paddings = zip(*[infer_single_layer_shape(sizes[i+1],sizes[i]) for i in reversed(range(len(sizes)-1))])
    max_pools = [1]*len(strides) # no max pool, just strides
    return EncByLayer(chans,ksizes,strides,paddings,max_pools,False)

def infer_single_layer_shape(in_size,out_size):
    stride_size = int(in_size/out_size)
    ksize = in_size - stride_size*(out_size - 1)
    tentative_out_size = (in_size - ksize)/stride_size + 1
    assert tentative_out_size == out_size
    padding = int((out_size - tentative_out_size)/2)
    if ksize<3:
        padding = int(math.ceil((3-ksize)/2))
        ksize += 2*padding
    tentative_out_size = (in_size + 2*padding - ksize)/stride_size + 1
    if not tentative_out_size == out_size:
        breakpoint()
    return ksize,stride_size,padding

if __name__ == '__main__':
    enc = build_conv_net(3,64,256,7)
    import pdb; pdb.set_trace()  # XXX BREAKPOINT
    print(enc(torch.ones(1,3,64,64)).shape)
    dec = build_convt_net(3,64,256,2)
    print(dec(torch.ones(1,256,1,1)).shape)